%% file: main.tex
\documentclass{egpubl}
\usepackage{egsgp2021}
\usepackage{booktabs, amsmath, makecell, amsfonts, dsfont}
\usepackage{xspace}
\usepackage{multirow, multicol}
\usepackage{subcaption}
\usepackage{titlesec}
\usepackage[absolute,overlay]{textpos}
 
\SpecialIssuePaper         % uncomment for final version of Computer Graphics Forum, special issue
\usepackage[T1]{fontenc}
\usepackage{dfadobe}  

\BibtexOrBiblatex
\electronicVersion
\PrintedOrElectronic

\ifpdf \usepackage[pdftex]{graphicx} \pdfcompresslevel=9
\else \usepackage[dvips]{graphicx} \fi

\usepackage{egweblnk} 
\usepackage{bbm}

\title[Roominoes: Generating Novel 3D Floor Plans From Existing 3D Rooms]
{Roominoes:\\Generating Novel 3D Floor Plans From Existing 3D Rooms}

\author[K. Wang et al.]
{\parbox{\textwidth}{\centering Kai Wang$^{1}$ \qquad
        Xianghao Xu$^{1}$                      \qquad
        Leon Lei$^{1}$                         \qquad
        Selena Ling$^{1}$                      \qquad
        Natalie Lindsay$^{1}$                  \\
        Angel X. Chang$^{2}$                   \qquad
        Manolis Savva$^{2}$                    \qquad
        Daniel Ritchie$^{1}$
        }
        \\
{\parbox{\textwidth}{\centering $^1$Brown University, United States\\
         $^2$Simon Fraser University, Canada
       }
}
}
\input{macros.tex}

\begin{document}

\teaser{
  \centering
  \setlength{\tabcolsep}{1pt}
  \vspace{-0.3em}
    \begin{tabular}{ccccccc}
        Relation Graph & 
        \multicolumn{5}{c}{\rule[1.5pt]{12em}{0.7pt} Iterative Room Retrieval \rule[1.5pt]{12em}{0.7pt}} &
        Final 3D Scene
        \\
        \raisebox{2em}{\multirow{2}{*}{\includegraphics[width=.13\linewidth]{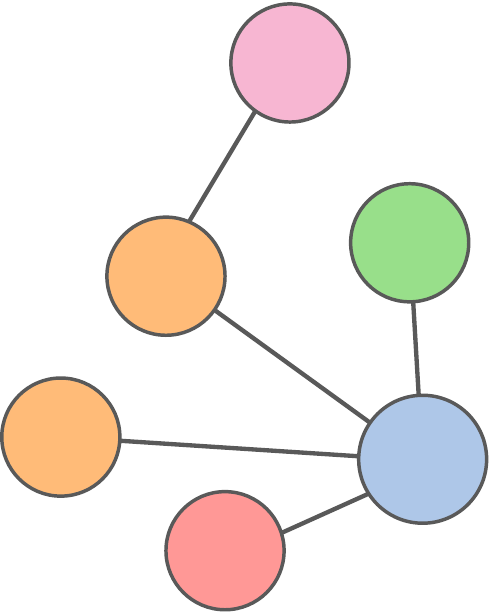}}} &
        \includegraphics[width=.12\linewidth]{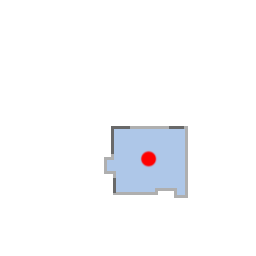} &
        \includegraphics[width=.12\linewidth]{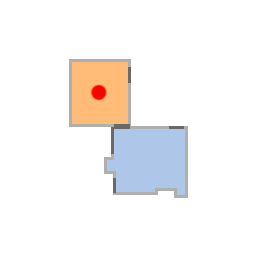} &
        \includegraphics[width=.12\linewidth]{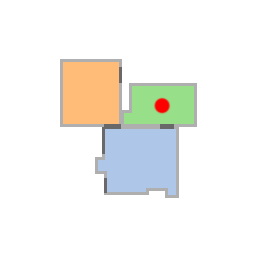} &
        \raisebox{3.5em}{\huge{$\cdots$}} &
        \includegraphics[width=.12\linewidth]{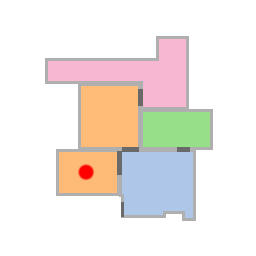} &
        \raisebox{2.5em}{\multirow{2}{*}{\includegraphics[width=.2\linewidth]{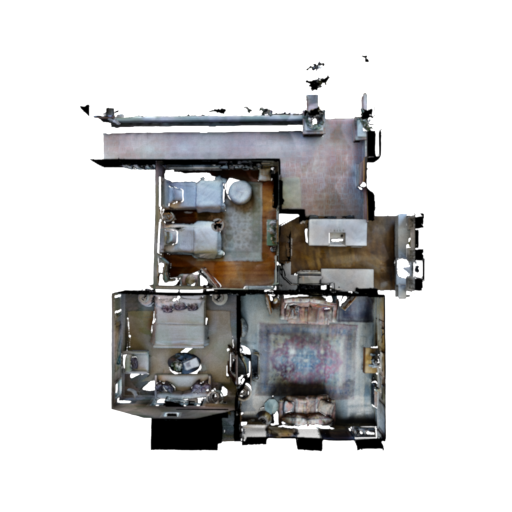}}} 
        \\
        & 
        \includegraphics[width=.10\linewidth]{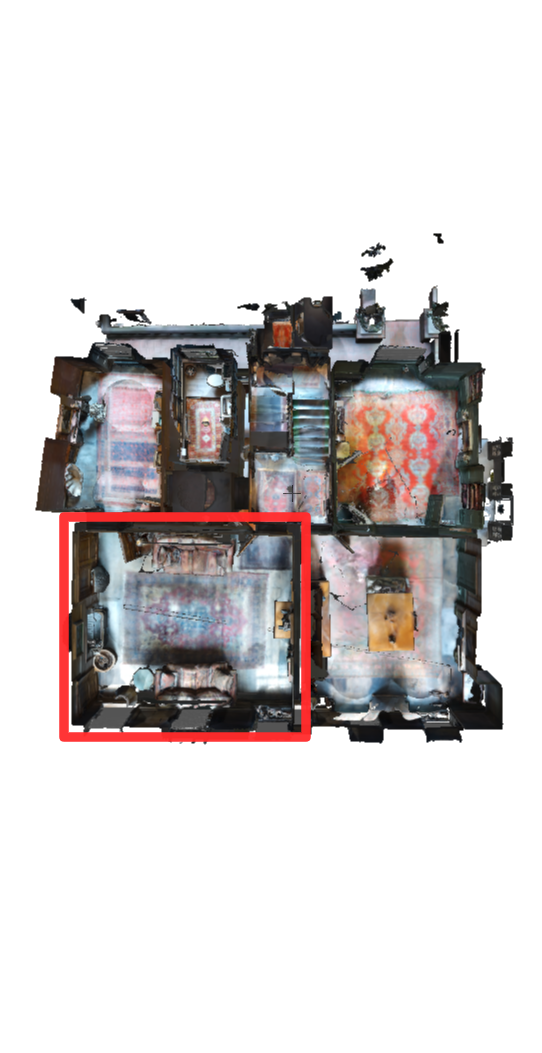} &
        \raisebox{3em}{\includegraphics[width=.09\linewidth]{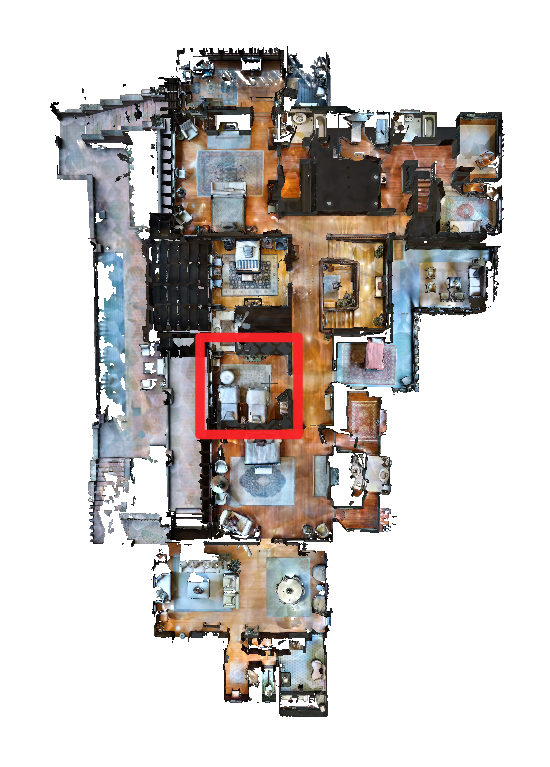}} &
        \raisebox{2.5em}{\includegraphics[width=.09\linewidth]{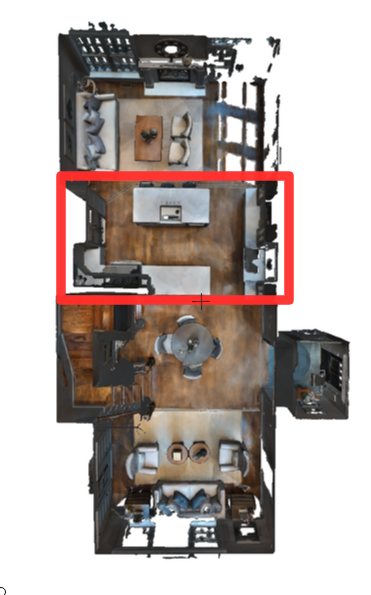}} &
        \raisebox{6em}{\huge{$\cdots$}} &
        \raisebox{2em}{\includegraphics[width=.09\linewidth]{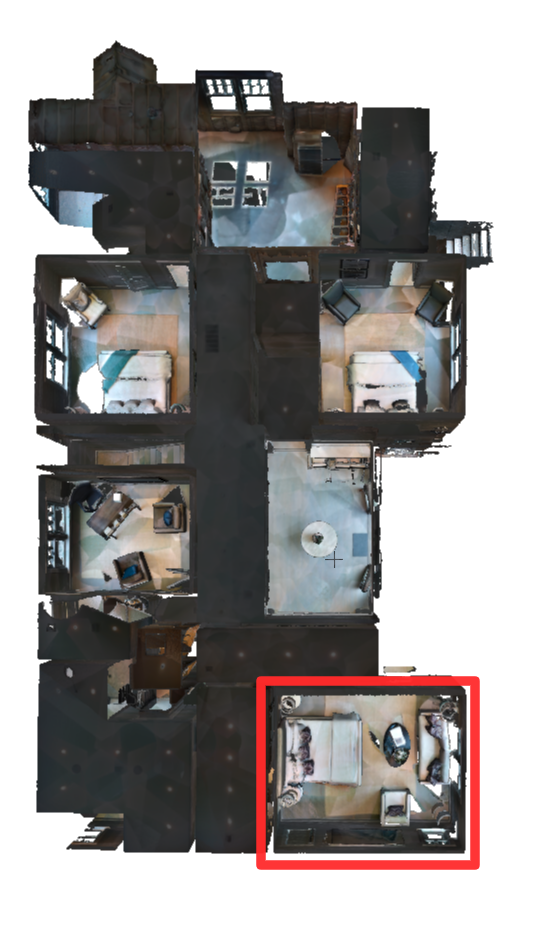}} &
    \end{tabular}
\vspace{-3.5em}
\caption{
We present Roominoes, a new task for creating house-level 3D environments by piecing together existing 3D rooms. 
We devise a spectrum of potential strategies to solve this task.
In an instance of this task, one of our proposed algorithms creates a combinatorially novel layout by iteratively retrieving and assembling rooms from different 3D floor plans, deforming each room as needed.
}
}

\maketitle

\begin{textblock*}{17.6cm}(1.75cm, 0.5cm)
\tiny \noindent
\emph{
This is the accepted version of the following article: Wang, K., Xu, X., Lei, L., Ling, S., Lindsay, N., Chang, A.X., Savva, M. and Ritchie, D. (2021), Roominoes: Generating Novel 3D Floor Plans From Existing 3D Rooms. Computer Graphics Forum, 40: 57-69, which has been published in final form at
https://onlinelibrary.wiley.com/doi/10.1111/cgf.14357.
This article may be used for non-commercial purposes in accordance with
the Wiley Self-Archiving Policy [http://olabout.wiley.com/WileyCDA/Section/id-820227.html].
}
\end{textblock*}
%-------------------------------------------------------------------------
\begin{abstract}
\input{sections/0-abstract.tex}
%-------------------------------------------------------------------------


\end{abstract}
%-------------------------------------------------------------------------
\input{sections/1-intro.tex}

\input{sections/2-relatedwork.tex}

\input{sections/3-overview.tex}

\input{sections/4-layoutfirst.tex}

\input{sections/5-retrievalfirst.tex}
\input{sections/6-deformation}
\input{sections/7-evaluation.tex}
\input{sections/8-conclusion.tex}
\input{sections/9-acknowledgements.tex}

%-------------------------------------------------------------------------
% bibtex
\bibliographystyle{eg-alpha-doi}  
\bibliography{main}

\setcounter{section}{0}
\renewcommand\thesection{\Alph{section}}
\input{supplementary_sections/1-optimization.tex}

\input{supplementary_sections/2-navigation.tex}
\input{supplementary_sections/3-walkthrough.tex}
\input{supplementary_sections/4-navigation.tex}
%-------------------------------------------------------------------------
%\newpage

\end{document}

%% file: macros.tex
\newcommand{\revisionnew}[1]{#1}
\newcommand{\revisiontwo}[1]{#1}

\newcommand{\indicator}{\ensuremath{\mathbbm{1}}}

\newcommand{\norm}[1]{\left\lVert#1\right\rVert}

\newcommand{\taskname}{Roominoes\xspace}

%% file: sections/0-abstract.tex
Realistic 3D indoor scene datasets have enabled significant recent progress in computer vision, scene understanding, autonomous navigation, and 3D reconstruction.
But the scale, diversity, and customizability of existing datasets is limited, and it is time-consuming and expensive to scan and annotate more.
Fortunately, combinatorics is on our side: there are enough individual \emph{rooms} in existing 3D scene datasets, if there was but a way to recombine them into new layouts.
In this paper, we propose the task of generating novel 3D floor plans from existing 3D rooms.
We identify three sub-tasks of this problem: generation of 2D layout, retrieval of compatible 3D rooms, and deformation of 3D rooms to fit the layout.
We then discuss different strategies for solving the problem, and design two representative pipelines: one uses available 2D floor plans to guide selection and deformation of 3D rooms; the other learns to retrieve a set of compatible 3D rooms and combine them into novel layouts.
We design a set of metrics that evaluate the generated results with respect to each of the three subtasks and show that different methods trade off performance on these subtasks.
Finally, we survey downstream tasks that benefit from generated 3D scenes and discuss strategies in selecting the methods most appropriate for the demands of these tasks.

%% file: sections/1-intro.tex
\section{Introduction}
\label{sec:intro}

Realistic 3D indoor scenes are useful for many tasks.
In recent years, datasets of 3D scenes have stimulated advances in computer vision~\cite{zhang2018deep,zamir2018taskonomy}, multimodal scene understanding~\cite{MatterportSim,chen2020soundspaces}, 3D reconstruction~\cite{LocalImplicitGrids,SceneCAD}, and visual navigation~\cite{gupta2017cognitive,GibsonEnv,AiHabitat}.
In particular, multi-room 3D interiors have enabled work on long-horizon embodied AI tasks involving both manipulation and navigation.

Unfortunately, the scale and diversity of existing 3D indoor scene datasets is a bottleneck for this research agenda: they are not very large nor configurable, and it is time-consuming and expensive to enlarge them.
For example, existing datasets provide mostly `detached' rooms (ScanNet~\cite{ScanNet}), or are limited to a small number of high-quality multi-room residences (Gibson~\cite{GibsonEnv}, Matterport3D~\cite{Matterport3D} and Replica~\cite{ReplicaDataset}).

\revisionnew{
Instead of using 3D reconstructions, one could turn to synthetic scene datasets, such as AI2-THOR~\cite{AI2Thor}, CHALET~\cite{CHALET} and 3D-FRONT~\cite{3DFront}.
One could also try to use generative models of furniture layouts to generate new synthetic 3D data~\cite{GRAINS, PlanIT, SceneGraphNet}.
Regardless of how synthetic scenes are created, there exists a ``reality gap'' between synthetic scenes and real ones, both in terms of content (e.g. synthetic scenes are not messy enough) and style (e.g. synthetic scenes are not photorealistic enough)~\cite{RenderingSUNCG}.
}

In this paper, we propose a new paradigm for creating house-level 3D environments by piecing together existing 3D rooms, a task that we call \emph{\taskname}.
This mix-and-match strategy is similar to room-level 3D scene synthesis that selects from a database of 3D objects and arranges them to create a furnished room.
Instead of arranging 3D objects, the \taskname task requires arranging entire rooms.
In other words, the rooms are building blocks to produce a coherent floor plan through a `retrieve and edit' approach.
This approach allows for scalability through combination of existing high-quality 3D rooms, as well as control of the level of complexity (from simple environments with few rooms, to complex environments with many rooms).

To analyze the challenges presented by the \taskname task, we decompose it into three sub-tasks: (1) generating a 2D floor plan that is compatible with available 3D rooms, (2) retrieving a set of 3D rooms that can fit into the 2D floor plan, and (3) deforming individual 3D rooms so they fit the 2D floor plan.
A \taskname method must address these three sub-tasks.
There are a spectrum of strategies for performing them, each with different tradeoffs.
One could use existing 2D floor plans to provide plausible layouts; however, fitting available 3D rooms into those layouts may induce undesirably large deformations to the 3D rooms and the objects they contain.
Alternatively, one could directly piece together existing 3D rooms without starting with a target layout; this will result in minimal distortion to the 3D rooms but may produce a less-realistic layout.
In this paper, we implement and evaluate these two strategies.
To measure the quality of the generated 3D environments, we introduce a set of evaluation metrics for measuring both the 2D layout quality and 3D mesh quality (e.g. amount of deformation, navigability).
We also discuss the potential usefulness of these generated environments for downstream tasks, and we demonstrate the applicability of generated 3D houses on a visual navigation task.

We find that the \taskname task is challenging, requiring non-trivial geometry processing to properly deform room architectures while keeping contained objects relatively undeformed.
We believe that compositional generation of realistic multi-room 3D houses will become increasingly useful.

%% file: sections/2-relatedwork.tex
\section{Background \& Related Work}
\label{sec:relatedwork}

\paragraph*{Multi-room 3D environment generation}
There is limited work that attempts to generate 3D environments with multiple rooms.  
Procedural content generation for studying RL is starting to leverage generated 3D environments.
However, these are mostly restricted to 3D mazes~\cite{beattie2016deepmind} or towers with rooms~\cite{juliani2019obstacle}, which do not reflect the complexity of real-world furnished houses.
A related line of work addresses procedural generation of floor plan layouts for residences and other buildings~\cite{ResidentialLayout,bao2013generating,LayoutExploration}.
These works, however, focus on generating the 3D architectural structure and do not typically produce realistically furnished 3D interiors.

\paragraph*{Modeling by Assembly}
Our task is related to the long line of research on 3D modeling by retrieving and assembling parts.
Early work in this area focused on interactive modeling, using either shape similarity search~\cite{ModelingByExample} or a learned probabilistic graphical model~\cite{SidVangelisAssembly} to retrieve relevant parts for a user to combine.
More recent approaches have automated both part retrieval and placement~\cite{SidVangelisSynthesis,ComplementMe}.
ComplementMe~\cite{ComplementMe} is most similar to one of our approach: it trains a deep metric learning model to retrieve 3D parts that are compatible with a partial query shape.
However, floor plans differ from part-based shapes both in data representation and in constraints that must be respected, which motivates us to develop a different formulation.

\paragraph*{Indoor Scene Synthesis}
Our problem statement is similar to that of indoor scene synthesis, which aims to generate room layouts comprised of existing 3D objects, whereas we focus on generating 3D layouts from existing 3D rooms.
One family of approaches to this problem attempts to place objects via optimization, typically a variant of Markov Chain Monte Carlo, with respect to learned priors, hand-crafted constraints, or some combination of the two~\cite{MakeItHome,InteractiveFurnitureLayout,SceneSynth,HumanCentricSUNCGSceneSynth}.
Another family of approaches trains deep generative models to output scenes, using either a latent variable model such as Variational Autoencoder~\cite{GRAINS} or Generative Adversarial Network~\cite{QixingSynth} or using an iterative object-by-object approach~\cite{DeepSynthSIGGRAPH2018,FastSynthCVPR, PlanIT}.
The iterative approaches are most similar to our setting.
However, floor plan generation is a considerably different domain than furniture layout, as inter-room relationships require more geometric precision than most inter-object relationships (e.g. walls must align, portals must connect).
Hence, a different approach is needed.

\paragraph*{Floor Plan Generation}
There is a large body of existing work on generating 2D floor plans.
As part of a larger system for generating 3D residential house models, Merrell et al. introduced a Bayesian network for modeling the room relationship graph along with an optimization-based approach for realizing this graph through precise wall geometry~\cite{ResidentialLayout}.
Liu et al. also use an optimization-based approach as part of a machine-assisted interactive system for designing precast concrete buildings~\cite{LayoutExploration}.
More recently, Wu et al. developed a mixed-integer quadratic programming (MIQP) formulation for optimizing building interior layouts~\cite{MIQPLayout}.
Finally, the past year has seen the introduction of deep learning methods into this problem space, resulting in learning-based methods for generating floor plans given building outlines~\cite{RPLAN}, room relationship graphs~\cite{HouseGAN}, or both~\cite{Graph2Plan}.
These methods all address generation of 2D layouts with unconstrained room shapes.
Very recent work takes a different approach of learning to generate a constraint graph and then solving the optimization problem posed by this graph to produce a final layout~\cite{GenerativeLayoutModelingConstraintGraphs}.
In contrast to all these approaches, we seek to solve the problem of generating new \emph{3D} layouts, where the room shapes are constrained by an available set of existing 3D rooms.

%% file: sections/3-overview.tex
\section{Overview}
\label{sec:overview}

Generating 3D floor plans using existing 3D rooms is a more constrained problem than typical 2D floor plan generation: one cannot generate rooms of arbitrary shapes, but instead is limited to the shapes available in an existing 3D room database.
It would be too restrictive, however, to allow no changes to the 3D room geometry at all, as it is unlikely that an existing set of 3D rooms can be recombined perfectly into a novel floor plan.
The likelihood of creating plausible layouts from existing rooms can be greatly increased if we allow deformations to the original rooms.
By distorting the outline of the room and moving the positions of connecting doors or spaces (which we call \emph{portals}), we can reuse the available 3D rooms in a more flexible manner and generate higher quality layouts.
However, this approach comes at the cost of degrading 3D room quality, which decreases as the amount of distortion increases.

There is no definitive answer on how to trade off between 2D layout plausibility and 3D mesh quality; different applications may warrant different tradeoffs.
Thus, we explore a spectrum of strategies which span this tradeoff space.
We propose to break down the \taskname task into three parts:
i) generating a 2D floor plan that is compatible with available 3D rooms;
ii) retrieving 3D rooms that can fit into the 2D floor plan; and
iii) deforming the 3D rooms so they better fit the 2D floor plan.
We note that the third step, room deformation, can be solved separately, provided that a correspondence can be found between each retrieved 3D room and its counterpart in the 2D floor plan.
Thus, our exploration of possible strategies focuses on approaches for solving the first two steps, taking into account the interaction between them.
In the remainder of this section, we first identify strategies for each of these steps and then propose two full pipelines, which we describe further in the following sections.

\subsection{Generating 2D Floor Plans}
3D interiors follow the structure of an underlying 2D floor plan.
\revisionnew{Taking into consideration that the rooms in a 2D floor plan must be matched with available 3D rooms, we propose the following strategies for obtaining a 2D layout:}

\paragraph*{\revisionnew{Existing Dataset}}
An obvious option for obtaining 2D floor plans is to draw them from an existing floor plan dataset, such as RPLAN~\cite{RPLAN} (80K annotated floor plans), or LiFULL~\cite{LiFULL} (millions of floor plan images).
These datasets are far larger than their 3D scene counterparts: the Matterport3D~\cite{Matterport3D} dataset, for example, contains only 184 floors. 
Using an existing floor plan guarantees the quality of the layout, but has other disadvantages.
The first issue is lack of control: if one wants to replace a bedroom with a living room, or to make a balcony larger, it is unlikely that another floor plan in the dataset satisfies such demands.
The second issue concerns the availability of 3D rooms which match the shape of the rooms in the 2D layouts.
In general, 3D rooms with exact shape matches are not available.
To fit the 3D rooms into the layout, large deformation is often necessary, which can impact the resulting 3D mesh quality.

\paragraph*{Generative Model}
Instead of using floor plans drawn directly from a dataset, one could also use one of the many available data-driven generative models of 2D floor plans to generate novel 2D layouts~\cite{RPLAN,Graph2Plan,HouseGAN}.
This approach will improve the control over the resulting layout, since these methods usually accept some user input specification (e.g. a room relationship graph) and are capable of generating multiple possible layouts given the same input specification.
However, this approach does not address the issue with 3D deformation artifacts.

\paragraph*{Naive Portal Stitching}
One can avoid these deformation artifacts entirely by ignoring 2D layout altogether: instead of combining 3D rooms according to a layout, we can instead stitch rooms together by connecting pairs of portals.
This approach removes the need to deform 3D rooms, but may produce an incoherent layout and is incompatible with downstream tasks that involve reasoning about the global scene layout.

\paragraph*{Smart Portal Stitching}
Finally, one can attempt to find a ``middle ground'' trade-off between 2D layout plausibility and amount of 3D deformation by modifying the portal stitching method above: identifying 3D rooms that can likely fit together into a plausible layout without causing too much deformation.

Table~\ref{tab:2D_strategies} summarizes the strengths and weaknesses of these different strategies for obtaining a 2D layout.

\begin{table}
    \centering
    \small
    \setlength{\tabcolsep}{4pt}
    \caption{
    Comparing different strategies for generating 2D layouts.
    }
    \begin{tabular}{@{}lccc@{}}
        \toprule 
        \textbf{Strategy} & \textbf{Layout Quality} & \textbf{Control} & \textbf{Deformation} \\
        \midrule
        \emph{Dataset} & High & Low & Large \\
        \emph{Generative Model} & High & Medium & Large \\
        \emph{Naive Portal Stitching} & Low & High & None \\
        \emph{Smart Portal Stitching} & Medium & High & Medium \\
        \bottomrule
    \end{tabular}
    \label{tab:2D_strategies}
\end{table}

\subsection{Retrieving Compatible 3D Rooms}

Depending on the strategy used to generate the 2D layout, the task of retrieving 3D rooms can be quite different.
If the entire 2D layout is known in advance, then the task becomes finding the 3D rooms that best match the rooms in the layout (in terms of shape and/or portal placements).
On the other hand, if the layout is determined on-the-fly (as in the two ``Portal Stitching'' approaches above), then the task becomes more ill-posed.
As there is no `ground truth' target room shape to match, one must instead retrieve rooms that are both geometrically compatible with the rooms already in the layout \emph{and} and lead to a globally plausible layout.

\subsection{Deforming 3D Rooms}

The final task involves taking the retrieved 3D rooms and deforming them according to the 2D floor plan.
The problem can be solved in two steps: first computing a correspondence between the original room and the target room, and then applying a standard mesh deformation algorithm according to the defined correspondence.
When the available 3D rooms come equipped with semantic object annotations, one can also leverage this information to avoid introducing visually objectionable artifacts such as non-rigid bending of semantically-meaningful rigid objects.

\subsection{Task Paradigms}

While the final task of deforming 3D rooms can be performed independently of the methods used for the first two tasks, these first two tasks are more correlated.
We identify two major paradigms for solving them together:
\begin{itemize}
    \item \textbf{2D before 3D:} Generate the 2D floor plan first and use its room shapes as ground truth shape targets for 3D room retrieval (i.e. the ``Dataset'' strategy in Table~\ref{tab:2D_strategies}).
    \item \textbf{2D by 3D:} Build up a layout iteratively by retrieving and connecting 3D rooms (i.e. ``Guided Portal Stitching'' in Table~\ref{tab:2D_strategies}).
\end{itemize}
In this paper, we propose one representative algorithm for each of the two paradigms. In Section \ref{sec:layoutfirst}, we describe a method that takes an existing 2D floor plan in the dataset, and retrieves 3D rooms similar according to the floor plan.
In Section \ref{sec:retrievalfirst}, we implement a method that uses deep metric learning to iteratively retrieve compatible 3D rooms and then apply \revisiontwo{a} \revisionnew{mixed integer quadratic programming (MIQP)} optimization to assemble them into the final layout. 
Finally, in Section \ref{sec:deformation}, we describes a strategy for deforming the 3D rooms, by first optimizing for a set of dense correspondences between the room outlines, and them applying cage deformation to deform the 3D rooms to the target outline.

\begin{figure*}
    \centering
    \setlength{\tabcolsep}{0pt}
    \renewcommand{\arraystretch}{0}
    \begin{tabular}{cccccccc}
        Query Room &
        \multicolumn{4}{c}{\rule[1.5pt]{8.5em}{0.7pt} Best 4 Retrievals \rule[1.5pt]{8.5em}{0.7pt}} & $10\%$ & $50\%$ & Worst
        \\
        \includegraphics[width=.12\linewidth]{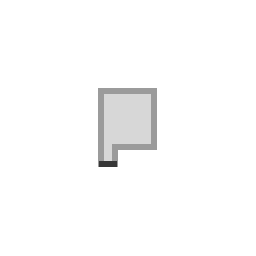} &
        \includegraphics[width=.12\linewidth]{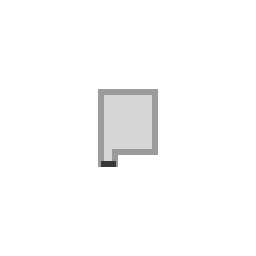} &
        \includegraphics[width=.12\linewidth]{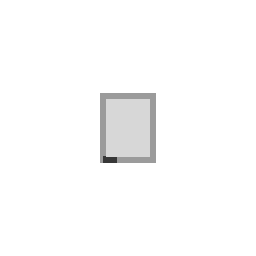} &
        \includegraphics[width=.12\linewidth]{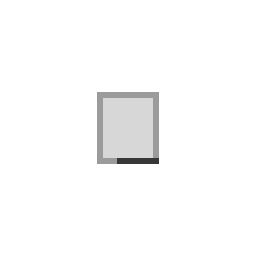} &
        \includegraphics[width=.12\linewidth]{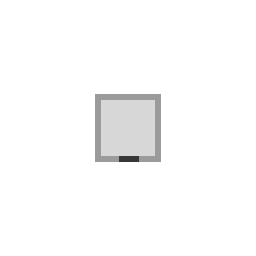} &
        \includegraphics[width=.12\linewidth]{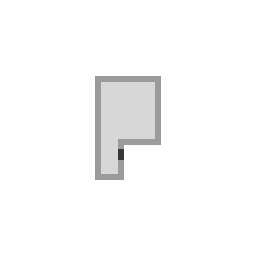} &
        \includegraphics[width=.12\linewidth]{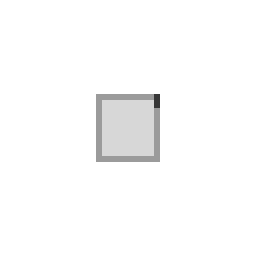} &
        \includegraphics[width=.12\linewidth]{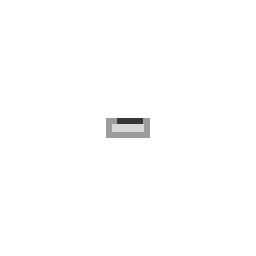} 
        \\[-2em]
        \includegraphics[width=.12\linewidth]{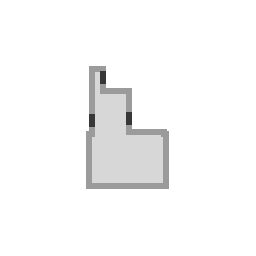} &
        \includegraphics[width=.12\linewidth]{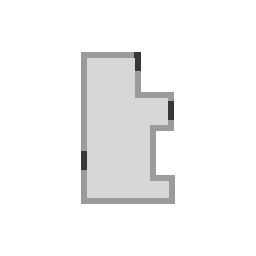} &
        \includegraphics[width=.12\linewidth]{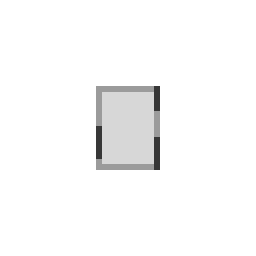} &
        \includegraphics[width=.12\linewidth]{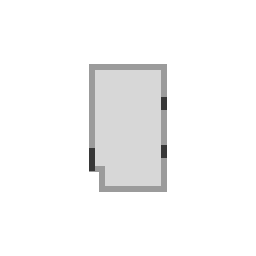} &
        \includegraphics[width=.12\linewidth]{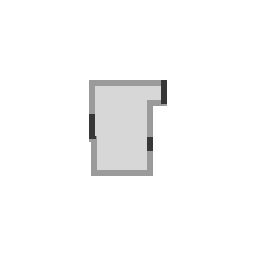} &
        \includegraphics[width=.12\linewidth]{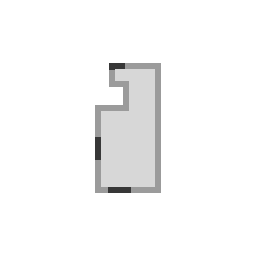} &
        \includegraphics[width=.12\linewidth]{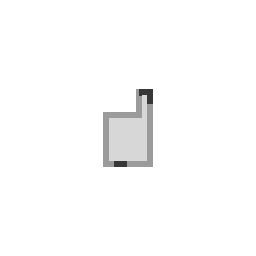} &
        \includegraphics[width=.12\linewidth]{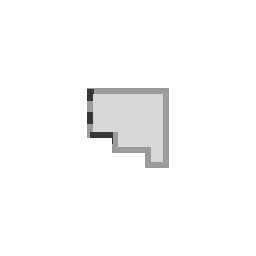}
        \\[-1.5em]
    \end{tabular}
    \caption{
    Given a 2D room in a floor plan, we find similar 3D rooms that can be deformed into the 2D room with minimal changes with regard to room size, shape, and portal locations. In the top row, the top retrievals are compatible with respect to all criteria, whereas subsequent retrievals start to different in terms of shape, portal location, and finally size. The bottom row shows that the difficulty of finding a compatible 3D room increases substantially when the room shape becomes more non-rectangular, and when the number of portals is greater than $1$.
    }
    \label{fig:room_match_rank}
\end{figure*}

%% file: sections/4-layoutfirst.tex
\section{Generating 2D Layout Before Retrieving 3D Rooms}
\label{sec:layoutfirst}

In this section, we describe an algorithm that utilizes a large collection of available 2D floor plans to guide the retrieval of 3D rooms.
Since the 2D layout is given by a floor plan from the dataset, the focus of this algorithm is on minimizing deformation of the retrieved 3D rooms needed to fit into the 2D floor plan.
Since it is computationally expensive to perform deformation for all available 3D rooms, we instead develop approximate metrics to assess how much deformation would be required.

\subsection{Data Preparation}
As our source of 2D floor plans, we use the RPLAN dataset~\cite{RPLAN}, a collection of $80,000$ floor plan images annotated at pixel level.
As in prior work, we apply additional filtering to the data: we make all walls the same thickness, retain only the largest portal between each pair of connecting rooms, and remove floor plans containing multi-purpose rooms. 

As our source of 3D room data, we use the Matterport3D dataset~\cite{Matterport3D}, a collection of $184$ floor plans containing $2056$ rooms.
For each 3D room, we build a 2D representation equivalent to those in the 2D floor plan dataset by extracting its outline (given by human annotations) and using raycasting to identify open regions on its walls, which we consider as portals. 
Since all of our 2D rooms are rectilinear, we also convert the non-rectilinear architecture of 3D rooms by computing their rectilinear bounding cages (i.e. the union of axis-aligned bounding boxes of all wall segments).
We discard rooms that are not closed as well as rooms where a portal falls on a non-rectilinear part of the wall. 
We also discard room types not present in RPLAN.
This leaves us with a final retrieval database of $1036$ rooms.
Finally, we apply a four way rotational augmentation of this dataset.

\subsection{Retrieving Compatible 3D Rooms}

Given a room from the 2D floor plan, we first filter out a subset of the 3D rooms that are compatible: those that have the same type label and contain same number of portals. 
For each 3D room in the filtered subset, we compute the following matching score between the source 3D room $R_S$ and the target 2D room $R_T$, each containing $k$ portals $P_S^{0} \ldots P_S^{k-1}$ and $P_T^{0} \ldots P_T^{k-1}$:
\[
\lambda_a c_{\text{area}} + \lambda_o c_{\text{outline}} + \lambda_p c_{\text{portal}}
\]
where 
\[
c_\text{area} = \frac{\max(A(R_S), A(R_T))}{\min(A(R_S), \revisionnew{A(R_T)})} - 1
\]
penalizes large difference in room areas $A(R_S), A(R_T)$, which will lead to large uniform scaling in the deformation process.
\[
c_\text{outline} = d_\text{chamfer}(O(R_S), O(R_T))
\]
computes the two-way chamfer distance between a $250$ point sample of the room outlines $O(R_S), O(R_T)$, with the rooms normalized to have unit area and centered. This penalizes room outline differences which lead to nonrigid deformations.
\[
c_\text{portal} = \min_{j=0}^{k-1}\sum_{i=0}^{k-1} c_\text{match}(P_S^i, P_T^{(i+j)\mathrm{mod}k})
\]
tries all possible $k$ ways of pairing the $k$ portals.
There are only $k$ ways because the order of the portals along the outline cannot be changed, and pairing any two portals uniquely determines the pairing of the rest.
The portal matching cost $c_\text{match}$ is:
\begin{align*}
    c_\text{match}(P_1, P_2) = \ &(M(P_1)-M(P_2))^2 + \\
    &\lambda_l (|P_1|-|P_2|)^2 + \\
    &\lambda_d \revisionnew{\mathds{1}_{\{d(P_1) \ne d(P_2)\}}}
\end{align*}
where $M(P)$ denotes the position of the midpoint of the portal in the 1D parametrized, unit length, outline of the room, $|P|$ denotes the length of the portal in the same 1D parametrization, and $\mathds{1}$ is the indicator function that takes the value of $1$ if the front-facing directions of the portals $d(P_1)$, $d(P_2)$ are different. This penalizes matching portals that might lead to large deformation: we don't want to slide the portal along the wall by too much, scale it too much, or put it in the wrong orientation.
We use $\lambda_a=1, \lambda_o = 0.5, \lambda_p = 10, \lambda_l = 0.5, \lambda_d = 0.05$ for all experiments.

Figure~\ref{fig:room_match_rank} shows examples of rooms retrieved using this score. The score allows retrieval of rooms that are similar regarding size, shape and portal locations. However, due to the limited availability of 3D rooms, it becomes increasingly unlikely to get an exact match as the number of portals and the complexity of the room shape increase.

\begin{figure*}
    \centering
        \includegraphics[width=\linewidth]{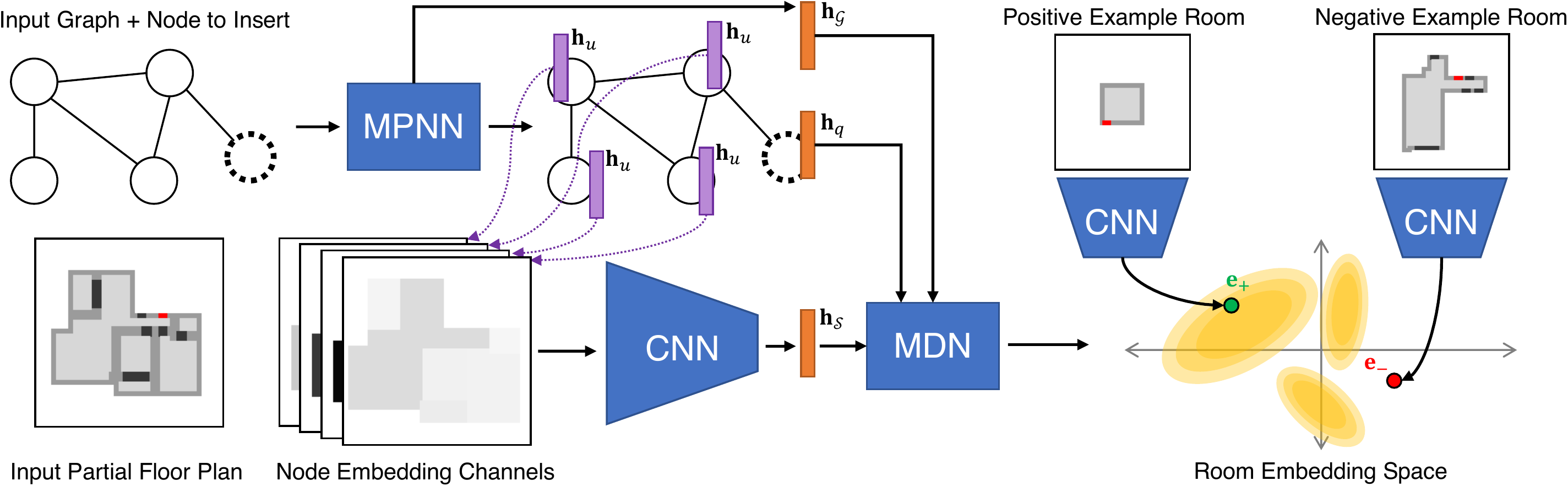}
    \caption{
    The architecture of our neural network-based room retrieval module.
    The \emph{embedding network} maps top down views of rooms into an embedding vector space (Right).
    Then, given an input floor plan relationship graph (with the node corresponding to the room to be inserted marked) and a top-down view of the current partial floor plan, the \emph{retrieval network} predicts a Gaussian mixture probability distribution over this embedding space.
    The two networks are trained jointly such that positive example rooms (obtained by removing random rooms from ground-truth floor plans) have higher probability than negative example rooms (random distractors).
    }
    \label{fig:retrieval_architecture}
\end{figure*}

%% file: sections/5-retrievalfirst.tex
\section{Generating 2D Layout By Retrieving 3D Rooms}
\label{sec:retrievalfirst}

While guiding 3D retrieval with existing 2D floor plans ensures the quality of the resulting scene's 2D layouts, it can lead to large deformations of 3D rooms.
To reduce the amount of deformation, it is possible to instead build the layout `on the fly': retrieving 3D rooms one by one and iteratively assembling them into a plausible 2D layout, changing the rooms slightly as needed.
In this section, we describe one pipeline that follows this paradigm. We extract a graph abstraction of an existing 2D floor plan, where each node \revisionnew{encodes the type of the room} and and each \revisiontwo{unlabeled} edge represents a portal connection. Conditioned on the graph, we iteratively retrieve compatible 3D rooms and use an optimization procedure to assemble them into a partial floor plan, repeat these steps until the layout is complete.

\subsection{Room Retrieval}
Each iteration of the algorithm must retrieve a 3D room corresponding to a node in the input layout graph.
The retrieved room needs to be compatible not only with the shapes of the other rooms retrieved so far, but also with the rooms retrieved later on, in order to form a coherent floor plan. 
Instead of trying to manually select features that lead to such compatibility,
we formulate the retrieval problem as a deep metric learning problem: we learn an embedding space of possible room shapes, where rooms which are compatible with the same layouts should be grouped together.
This is similar in spirit to the approach taken by Sung et al. for learning to assemble part-based shapes~\cite{ComplementMe}.
Our neural architecture has two sub-networks: an \emph{embedding network} and a \emph{retrieval network}.
The embedding network maps individual 3D rooms into a high-dimensional vector space; the retrieval network maps an input partial floor plan, as well as a relation graph of the entire floor plan, into a probability distribution over this space.
The two networks are trained jointly using a contrastive learning framework, i.e. rooms that are known to fit well with a layout in the training set should map to higher-probability points than other rooms.
Figure~\ref{fig:retrieval_architecture} shows the architecture of these networks and their interaction; the rest of this section describes them in more detail.

\paragraph*{Embedding Network}
The embedding network maps a 2D room $r$ into an embedding vector $\mathbf{e}_r$.
For this, we pass an image-based representation of the room through a convolutional neural network (CNN).
The image representation uses separate channels to represent the rooms, walls, portals, as well as the specific portal to which we want to connect the next room.

\paragraph*{Retrieval Network}
The retrieval network takes the current partial floor plan $\mathcal{S}$, the target floor plan relation graph $\mathcal{G}$, and the node $q$ corresponding to the room to be inserted, and predicts a conditional probability distribution $P(\mathbf{e}_r\;|\; q, \mathcal{G},\mathcal{S})$ over room embeddings $\mathbf{e}_r$.
Similarly to the work of Sung et al.~\shortcite{ComplementMe}, we model this distribution as a mixture of Gaussians over a room shape embedding space.
We use a mixture density network~\cite{MDN} to predict this distribution.

Since the features of $\mathcal{G}$ and $\mathcal{S}$ are highly correlated, we use a single neural network to learn from both simultaneously.
We first use a message passing neural network~\cite{NeuralMessagePassing} to extract information from the relationship graph $\mathcal{G}$.
After $T$ rounds of message propagation, we predict a per-node descriptor $\mathbf{h}_u$, as well as a descriptor of the entire graph $\mathbf{h}_\mathcal{G}$.
To extract information from the partial floor plan, we use a CNN conditioned on an image-based representation of the floor plan to predict an image feature $\mathbf{h}_\mathcal{S}$.
The image-based representation is the same as that used by the embedding network, with an additional set of $|\mathbf{h}_u|$ channels.
These channels encode the corresponding node embedding $\mathbf{h}_u$ for each room in the layout.
Finally, we concatenate the image feature $\mathbf{h}_\mathcal{S}$, the graph descriptor $\mathbf{h}_\mathcal{G}$, as well as the descriptor for the room to be retrieved $\mathbf{h}_q$, and feed it to a mixture density network, which predicts the parameters of the Gaussian mixture, where the $k$-th Gaussian is parameterized by a weight $\phi_k$, a mean $\mu_k$ and a standard deviation $\sigma_k$.

\paragraph*{Joint Training}
We train the embedding and retrieval networks jointly using a triplet loss.
Given a room embedding $\mathbf{e}_r$, we define the log likelihood that the embedding is sampled from the mixture distribution predicted by the retrieval network:
\begin{equation*}
    \ell^{q, \mathcal{G},\mathcal{S}}(\mathbf{e}_r) = \log\sum_{k=1}^N \phi_k(q, \mathcal{G},\mathcal{S}) \cdot \mathcal{N} (\mathbf{e}_r \;|\; \mu_k(q, \mathcal{G},\mathcal{S}), \sigma_k(q, \mathcal{G},\mathcal{S})^2)
\end{equation*}
Using this embedding loss, we define a contrastive loss~\cite{TripletLoss} between an embedding of a positive example room $\mathbf{e}_{+}$ and embedding of a negative example room $\mathbf{e}_{-}$:
\begin{equation*}
    \mathcal{L}^{q, \mathcal{G},\mathcal{S}}(\mathbf{e}_{+}, \mathbf{e}_{-}) 
    = \max\{m + \ell^{q, \mathcal{G},\mathcal{S}}(\mathbf{e}_{-}) - \ell^{q, \mathcal{G},\mathcal{S}}(\mathbf{e}_{+}), 0\}
\end{equation*}
where $m=10$ is a constant margin.
In other words, this loss encourages positive example rooms to have a higher predicted likelihood than negative example rooms.
Positive example rooms are created by removing a room from one of the training layouts.
Negative examples are created by sampling rooms from other floor plans.
For rooms with more than one portal, we randomly select a subset of portals to show in the query portal mask.

\begin{figure*}[t!]
    \centering
    \setlength{\tabcolsep}{0pt}
    \renewcommand{\arraystretch}{0}
    \begin{tabular}{cccccccc}
        Graph Context &
        Floor Plan &
        \multicolumn{3}{c}{\rule[1.5pt]{6em}{0.7pt} Best 3 Retrievals \rule[1.5pt]{6em}{0.7pt}} & $10\%$ & $50\%$ & Worst
        \\
        \includegraphics[width=.12\linewidth]{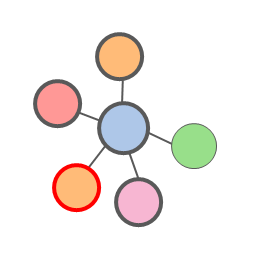} &
        \includegraphics[width=.12\linewidth]{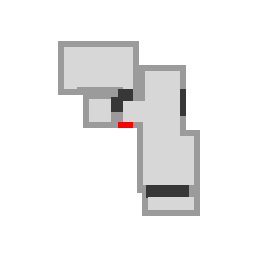} &
        \includegraphics[width=.12\linewidth]{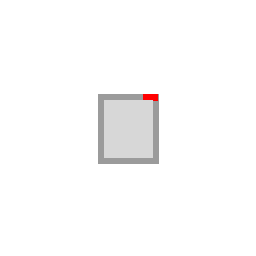} &
        \includegraphics[width=.12\linewidth]{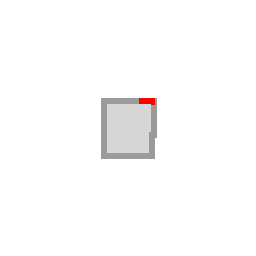} &
        \includegraphics[width=.12\linewidth]{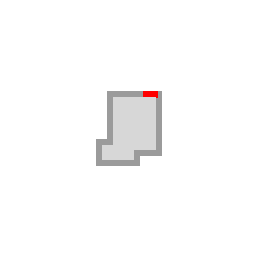} &
        \includegraphics[width=.12\linewidth]{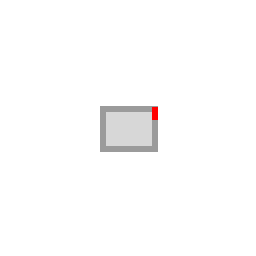} &
        \includegraphics[width=.12\linewidth]{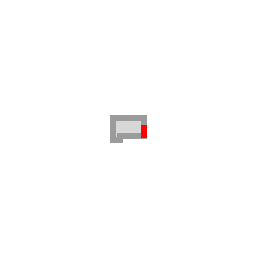} &
        \includegraphics[width=.12\linewidth]{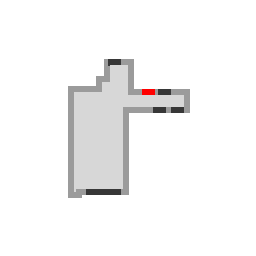}
        \\[-0.5em]
        \includegraphics[width=.12\linewidth]{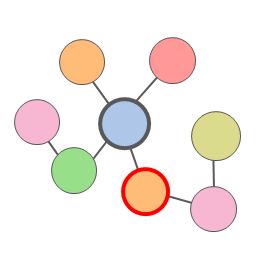} &
        \includegraphics[width=.12\linewidth]{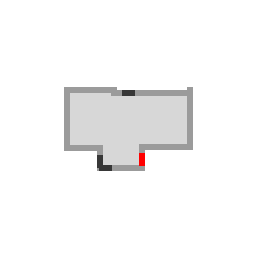} &
        \includegraphics[width=.12\linewidth]{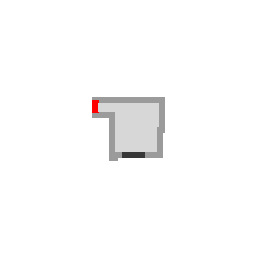} &
        \includegraphics[width=.12\linewidth]{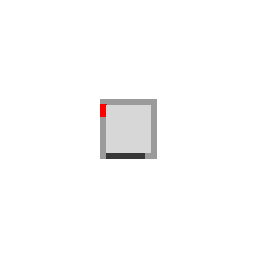} &
        \includegraphics[width=.12\linewidth]{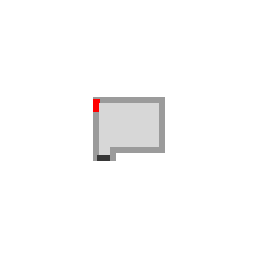} &
        \includegraphics[width=.12\linewidth]{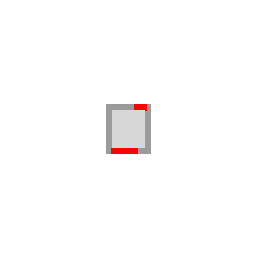} &
        \includegraphics[width=.12\linewidth]{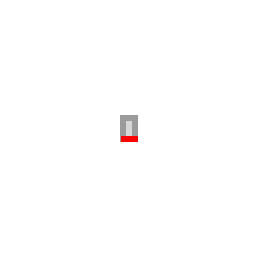} &
        \includegraphics[width=.12\linewidth]{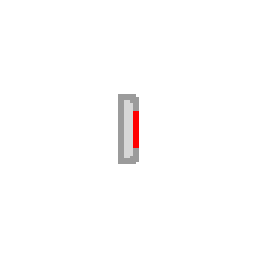}
        \\[-0.5em]
    \end{tabular}
    \caption{
    Our learned retrieval network retrieves rooms to add to an in-progress layout.
    The three highest-probability retrievals are good matches and permit the rest of the graph to be completed later.
    In the top row, the top 3 retrievals are all medium-sized rooms that can be easily inserted into the available corner given their portal placements.
    In the bottom row, the query node in the graph specifies that the new room should have two portals, and all the top 3 retrieval results do.
    The quality of retrieval falls off further down the probability-ordered list of rooms in our validation set, exhibiting errors such as incorrect room types, incorrect shapes, and incorrect numbers of portals.
    }
    \label{fig:retrieved_rooms}
\end{figure*}

We use $T=5$ rounds of propagation for the graph neural network.
We choose $|\mathbf{h}_u| = 64$ as the dimension of the per node embedding and $\mathbf{h}_\mathcal{G} = 128$ as the dimension of the graph embedding. 
The CNN contains seven layers, without batch normalization, which we find to be detrimental to the training process.
We use $|\mathbf{h}_\mathcal{S}| = 512$ as the dimension of predicted image feature.
For the retrieval network, we predict a mixture of $20$ Gaussians over an $128$ dimensional embedding space.
We use a margin of $10$ for the triplet contrastive loss.
We train the neural networks using the Adam~\cite{Adam} optimizer, and with a batch size of $16$. At each batch, negative examples are sampled from $16$ randomly selected floor plans. We start with uniformly selecting from these examples, gradually favoring harder examples as the training proceeds.

Figure~\ref{fig:retrieved_rooms} shows examples of rooms predicted by the retrieval network.
The network learns to retrieve rooms of the correct shape, correct type, and the correct number of portals.
\revisionnew{Note that the networks are not trained to recognize the possibility of using a room after rotations. To allow for this, we apply a four way rotational augmentation to all the candidate 3D rooms instead.}

\subsection{Placing Retrieved Rooms into the Floor Plan}

We use the trained embedding and retrieval networks to iteratively retrieve and insert rooms into the layout.
We determine the location of the new room by matching the position of the portal to the connecting rooms.
When multiple such portals exist, we randomly select one for initialization.

\begin{figure}[t!]
    \centering
    \setlength{\tabcolsep}{0pt}
    \renewcommand{\arraystretch}{0}
    \begin{tabular}{cccc}
        Retrieval & Optimization & Retrieval & Optimization 
        \\
        \includegraphics[width=0.24\linewidth]{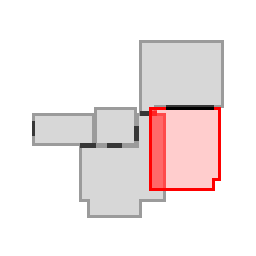} &
        \includegraphics[width=0.24\linewidth]{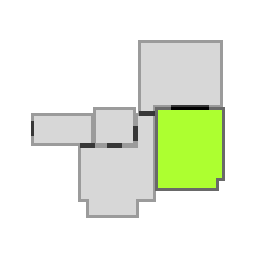} &
        \includegraphics[width=0.24\linewidth]{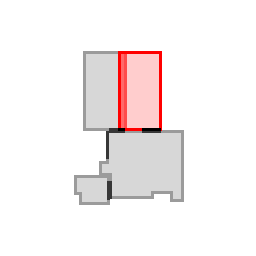} &
        \includegraphics[width=0.24\linewidth]{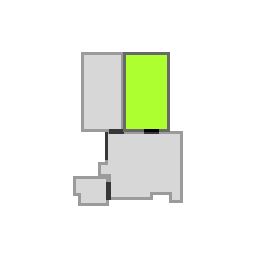}
    \end{tabular}
    \caption{Examples of a 2D floor plan before and after our optimization-based snapping.
    From left to right: fixing a room slightly too large to fit in a small corner; sliding a pair of portals to make the new room match better.
    }
    \label{fig:optimize_layout_result}
\end{figure}

\begin{figure}[t]
    \centering
    \setlength{\tabcolsep}{0pt}
    \renewcommand{\arraystretch}{0}
    \begin{tabular}{cccc}
         Without & With & Without & With
         \\
         \includegraphics[width=0.23\linewidth]{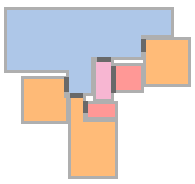} &
         \includegraphics[width=0.23\linewidth]{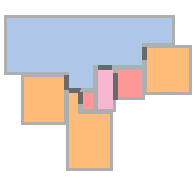} &
         \includegraphics[width=0.23\linewidth]{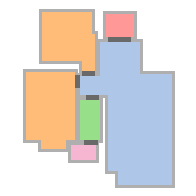} &
         \includegraphics[width=0.23\linewidth]{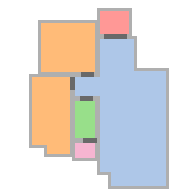}
    \end{tabular}
    \caption{
    Two examples illustrating the effect of rewarding the layout optimizer for maximizing adjacencies between different rooms.
    }
    \label{fig:adjacency_reward}
\end{figure}

However, it is unlikely that the retrieved room will fit perfectly into the current partial floor plan.
We use beam search to increase the chance that we find better fits.
To resolve remaining inconsistencies, we apply an optimization procedure after each step of room retrieval. 
Inspired by prior work~\cite{MIQPLayout,GenerativeLayoutModelingConstraintGraphs}, we adopt a mixed integer quadratic programming (MIQP) based procedure for layout optimization.
The optimization procedure has two goals.
The first is to ensure that the partial floor plan is valid, i.e. each room has a positive area, no overlaps exist between different rooms, and all paired portals align with each other.
Figure~\ref{fig:optimize_layout_result} shows examples of this process.
The second goal is to avoid obvious semantic issues in the floor plan.
We tackle one such issue, interior voids in the floor plan, by encouraging the rooms to be adjacent to each other, \revisionnew{though additional heuristics could be employed to improve the quality of the 2D floor plans further.}
Figure~\ref{fig:adjacency_reward} shows the effect of \revisionnew{encouraging room adjacency}.
These goals must be achieved while minimizing deformation of the room outlines and the portal locations.
The precise formulation of the objective function and constraints can be found in the supplemental material.

%% file: sections/6-deformation.tex
\section{Deforming Retrieved 3D Rooms}
\label{sec:deformation}

Having generated a 2D layout and retrieved 3D rooms, the final task involves deforming the individual room geometries to fit the layout.
This is a nontrivial problem: the 3D room must be warped to fit the new 2D outline, but we must take care to avoid introducing visually-objectionable artifacts, most prominently non-rigid distortion to semantically-meaningful objects within the room.

Deforming a 3D room to fit an optimized floor plan layout amounts to solving an analogy problem: given the source 3D room mesh $\mathcal{M}_S$ and its 2D outline polygon $\mathcal{P}_S$, as well as the target 2D outline polygon $\mathcal{P}_T$ produced by the optimizer, what is the corresponding target 3D room mesh $\mathcal{M}_T$?

In this section, we describe a two-step solution.
First, we warp the source outline $\mathcal{P}_S$ onto the target outline $\mathcal{P}_T$ by optimizing for a dense correspondence between them.
Then, we use the warped outline as a control cage for cage-based deformation of the source mesh $\mathcal{M}_S$ to produce the target mesh $\mathcal{M}_T$.

\paragraph*{Finding a correspondence between outlines}
Our first step in deforming the 3D room mesh to fit the new target room outline $\mathcal{P}_T$ is to first deform its 2D outline $\mathcal{P}_S$ to the target outline.
As both outlines are closed, one-dimensional, piecewise linear curves, this problem reduces to finding a correspondence between $\mathcal{P}_S$ and $\mathcal{P}_T$.
We solve this problem by sampling a finite set of $N$ 1D points $u^0_S \ldots u^{N-1}_S$ along the source outline $\mathcal{P}_S$ (including all of its corner points) and then optimizing for a corresponding set of 1D points $u^0_T \ldots u^{N-1}_T$ on $\mathcal{P}_T$, by minimizing the following objective:
\begin{align*}
    \sum_{i=0}^{N-1} \frac{\lambda_e}{N} c_{\text{elasticity}}(i) + \frac{\lambda_n}{N} c_{\text{normal}}(i) - r_{\text{corner}}(i)
\end{align*}
where 
\begin{equation*}
    c_\text{elasticity}(i) = \bigg((u^{(i+1)\mathrm{mod} N}_T - u^{i}_T) - \frac{|P_T|}{N} \bigg) ^2
\end{equation*}
is the elasticity regularization term that encourages the output points $u^i_T$ to be spread out evenly on the target outline, 
\begin{equation*}
    c_\text{normal}(i) = \theta_i\bigg(x_T^{(i+1)\mathrm{mod} N} - x_T^{i}\bigg)^2 + (1-\theta_i)\bigg(y_T^{(i+1)\mathrm{mod} N} - y_T^{i}\bigg)^2
\end{equation*}
is the normal matching term that encourages originally vertical segments to stay vertical ($\theta_i = 1$) and originally horizontal segments to stay horizontal ($\theta_i = 0$), where $(x^i_T, y^i_T)$ is the 2D position of $u^i_T$ along $\mathcal{P}_T$, and 
$r_\text{corner}(i) = \sigma_S^{i} \sigma_T^{i}$
is the term that encourages source corner points to correspond to target corner points ($\sigma^i \in \{0,1\}$ is a constant indicating whether the point $i$ is a corner in the source/target).

In addition, we impose the constraint that $u_T^i < u_T^{i+1}$ for all points $i$, which enforces that the sequence of points remains monotonic.
To make sure that the portals fall on the right segment of the outline after deformation, we impose the additional constraint that for each pairs of portals $p_S^i$ and $p_T^i$, the endpoints of the original portal $\mu_S^{ia}$ and $\mu_S^{ib}$ falls within the corner points $\mu_T^{ia}$ and $\mu_T^{ib}$ that contain the target portal.
For both the source and target outlines, we define $u = 0$ to occur at the upper-left-most corner of the outline.

\begin{figure}
    \centering
    \setlength{\tabcolsep}{1pt}
    \begin{tabular}{cccc}
        Input Room & Input Outline & Target Outline & Deformed Room
        \\
        \includegraphics[width=.24\linewidth]{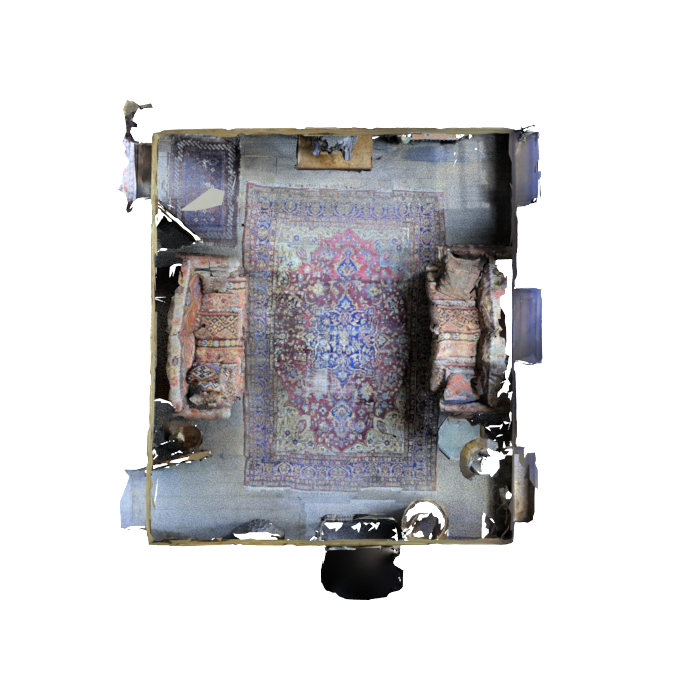} &
        \includegraphics[width=.24\linewidth]{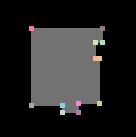} &
        \includegraphics[width=.24\linewidth]{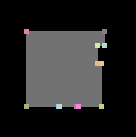} &
        \includegraphics[width=.24\linewidth]{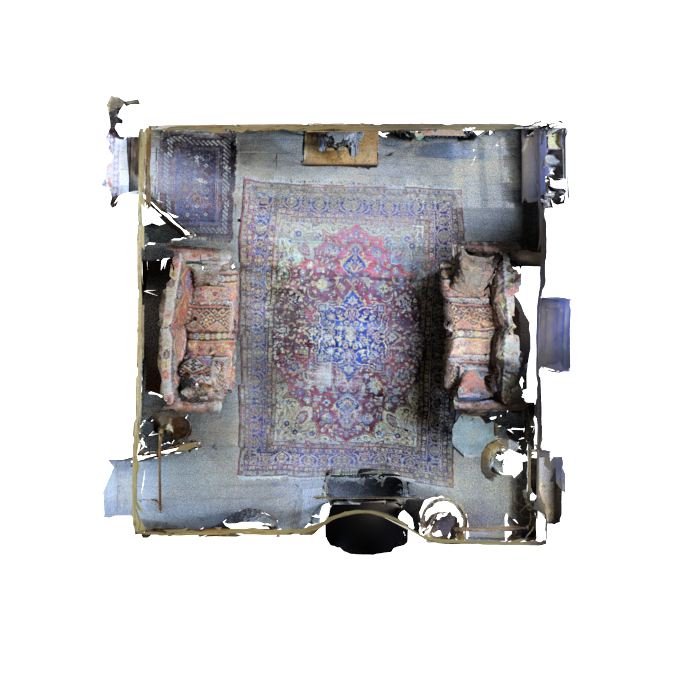}
    \end{tabular}
    \caption{
    Example of a 3D room deformed to fit an optimized 2D outline.
    From left to right: the initial 3D room mesh; the 2D outline for the input mesh (with a subset of sample points $u^0_S \ldots u^{N-1}_S$  colored); the target 2D outline for the deformation (with corresponding points $u^0_T \ldots u^{N-1}_T$ colored ); the final deformed room geometry.
    }
    \label{fig:3D_transfer}
\end{figure}

\paragraph*{Outline-driven deformation}
Given the warped outline produced by the correspondence step, we drive a cage-based deformation of the room mesh.
That is, $\mathcal{P}_S(u^0_S \ldots u^{N-1}_S)$ is treated as the initial polygonal control cage for the mesh, and $\mathcal{P}_T(u^0_T \ldots u^{N-1}_T)$ provides the new positions of the cage vertices.
We then interpolate the positions of all interior mesh vertices using mean value coordinates (MVC)~\cite{MeanValueCoordinates}.
Finally, we deform each wall where a portal falls on to make sure the portals fall on the right position.
Figure~\ref{fig:3D_transfer} shows an example of these procedures applied to a 3D room.

\begin{figure}
    \centering
    \setlength{\tabcolsep}{1pt}
    \begin{tabular}{ccc}
        \makecell{Input Room} & w/out Rigid Objects & w/ Rigid Objects
        \\
        \includegraphics[width=.32\linewidth]{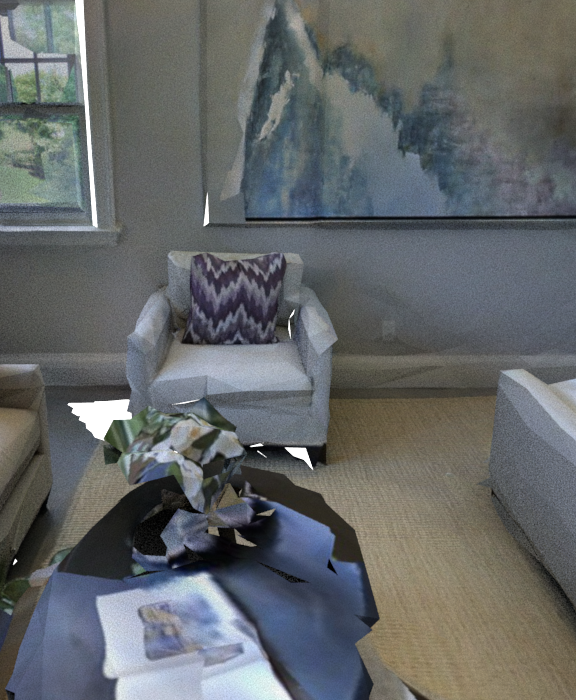} &
        \includegraphics[width=.32\linewidth]{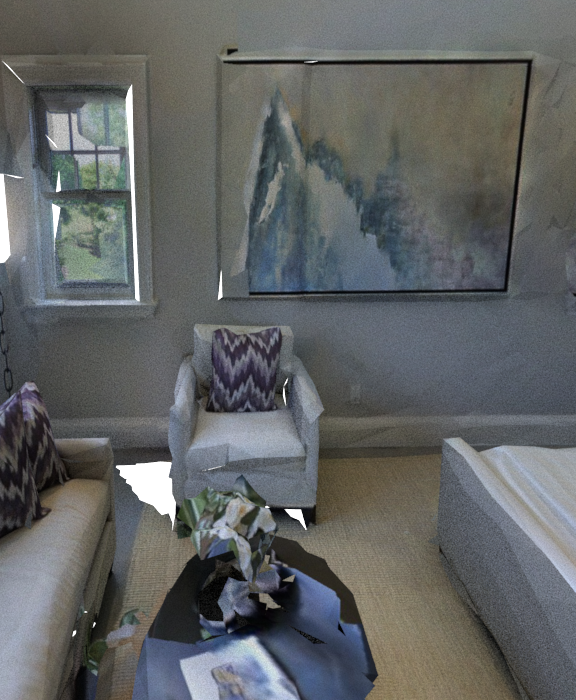} &
        \includegraphics[width=.32\linewidth]{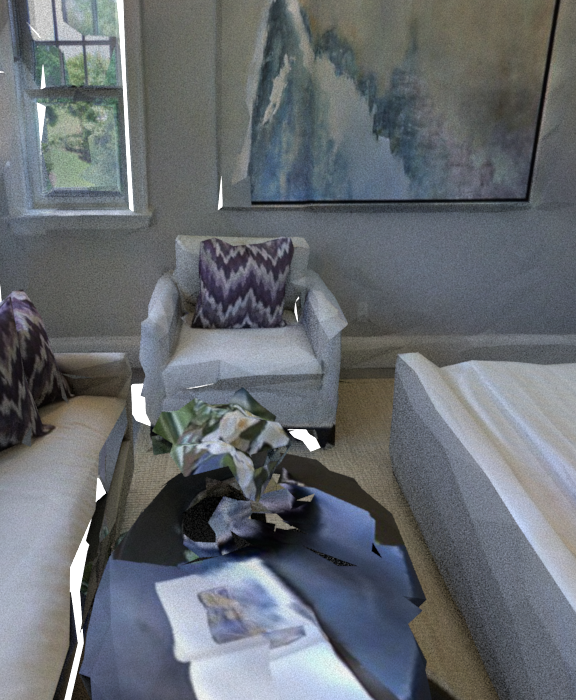}
    \end{tabular}
    \caption{
    Examining the effect of treating objects as rigid in our 3D room deformation procedure.
    From left to right: a view of the initial 3D room mesh; the room deformed without treating objects as rigid; the room deformed while treating objects as rigid.
    Enforcing object rigidity prevents semantically implausible bending artifacts.
    }
    \label{fig:3D_transfer_ablation}
\end{figure}

\paragraph*{Handling rigid objects}
The deformation scheme described above non-rigidly distorts the room geometry.
Since room retrieval and layout optimization try to retrieve and place rooms that fit into a layout together without changing their shape much, this distortion is typically small and unobjectionable.
However, when objects within a room undergo such deformations, it can result in semantically-implausible bending artifacts.
To prevent such artifacts, we cut \revisiontwo{all labeled objects other than those with labels floor, ceiling, wall or curtain} out of the mesh, deform the rest of the mesh using the scheme above, and then insert the objects back into the deformed room mesh.
To determine object insertion positions, we move their original centroid position according to the MVC deformation.
In some cases, this results in placing the object in a position that intersects with other objects or the room geometry.
If this occurs, we push the object away from the collision until the intersection is resolved.
If doing so results in another intersection before the first is resolved, we uniformly scale the object down until the collision is resolved.
Figure~\ref{fig:3D_transfer_ablation} shows deformation results with and without this special logic for handling rigid objects.

%% file: sections/7-evaluation.tex
\section{Evaluation Strategies for 3D Layout Generation}
\label{sec:evaluation}

In this section, we propose a set of metrics for evaluating the effectiveness of methods that solve the Roominoes task.
First, we identify the properties of the output that are of interest to downstream tasks.
We then define metrics that estimate the quality of the results, without having to spend many machine hours to actually test the generated data on downstream tasks.
Finally, we demonstrate that 3D scenes generated by \taskname can be useful for downstream tasks.

\subsection{Evaluating 2D Layout Quality}

The quality of a generated floor plan's 2D layout is important for tasks that require reasoning about the global structure of a scene, e.g. indoor navigation.
We propose the following metrics to evaluate 2D layout quality:
\begin{itemize}
    \item \textbf{Fr\'echet Distance (FD)}: a measure of distributional similarity between the generated floor plans and those in a held-out test set~\cite{FrechetInceptionDistance}.
    We evaluate this distance in two different feature spaces: a pre-trained Inception image classifier, i.e. the standard Fr\'echet Inception Distance (\textbf{FID}), and the CNN component of the retrieval network described in Section~\ref{sec:retrievalfirst} (\textbf{FD[R]}). The first is more general, whereas the second is domain-specific.
    \item \textbf{Classifier Fool Percentage}: the percentage of held-out test layouts classified as ``generated'' by a classifier trained to distinguish between generated 2D layouts and layouts from the training dataset. A value of $50\%$ would suggest that the two sets of layouts are indistinguishable from each other.
\end{itemize}

\noindent
We compare the following sources of 2D floor plans:
\begin{itemize}
    \item \textbf{Rectangles:} Randomly place $N$ rectangles, where $N$ equals the number of nodes in the graph. This simple baseline establishes a lower bound on performance.
    \item \textbf{Dataset:} Floor plans drawn from a dataset (RPLAN~\cite{RPLAN}) that are not used for training the following models.
    \item \textbf{Generative Model:} Generate floor plan given the polygonal outline of the building, using a recent method~\cite{RPLAN}.
    \item \textbf{Smart Portal Stitching}: Build a 2D layout while retrieving 3D rooms using the algorithm described in Section~\ref{sec:retrievalfirst}.
    \item \textbf{Smart Portal Stitching (no net)}: Use the algorithm from Section~\ref{sec:retrievalfirst}, but do not use the learned retrieval network; instead, retrieve random rooms with correct type and portal number.
\end{itemize}

\begin{table}
    \centering
    \setlength{\tabcolsep}{3pt}
    \caption{
    Comparing different methods for 2D floor plan generation.
    }
    \begin{tabular}{@{}lccc@{}}
        \toprule
        \textbf{Method} & \textbf{FID} $\Downarrow$ & \textbf{FD[R]} $\Downarrow$ & \textbf{\% fool} $\Uparrow$ \\
        \midrule
        Rectangles & $63.49$ & $450.74$ & $1.34$ \\
        Dataset & $6.84$ & $1.29$ & $53.54$ \\
        Generative Model & $9.36$ & $2.72$ & $46.86$ \\
        Smart Portal Stitching & $18.61$ & $86.05$ & $28.72$ \\
        Smart Portal Stitching (no net) & $22.40$ & $90.20$ & $25.00$ \\
        \bottomrule
    \end{tabular}
    \label{tab:2d_comparisons}
\end{table}

\begin{figure*}
    \centering
    \setlength{\tabcolsep}{0pt}
    \renewcommand{\arraystretch}{0}
    \begin{tabular}{rcccccc}
          \\[1.5em]
          \multicolumn{7}{c} {Match 2D Layout Shape (2D: Dataset)}
          \\[0.5em]
          \raisebox{2.8em}{\rotatebox{90}{2D Plan}} &
          \includegraphics[width=0.16\linewidth]{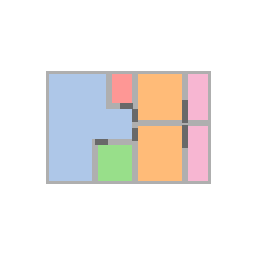} &
          \includegraphics[width=0.16\linewidth]{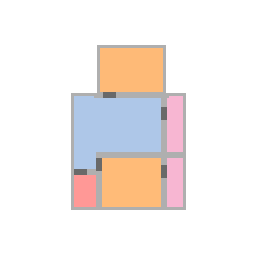} &
          \includegraphics[width=0.16\linewidth]{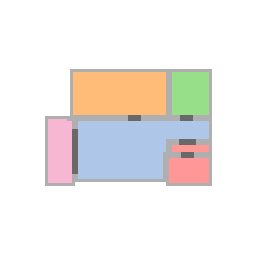} &
          \includegraphics[width=0.16\linewidth]{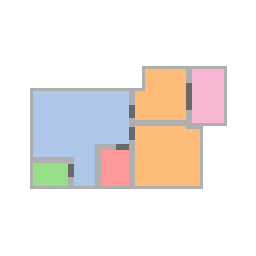} &
          \includegraphics[width=0.16\linewidth]{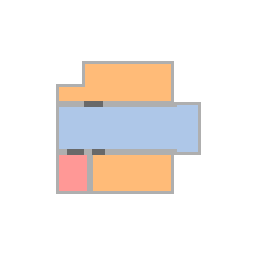} &
          \includegraphics[width=0.16\linewidth]{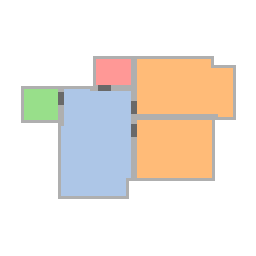} \\[-0.5em]
          \raisebox{2.25em}{\rotatebox{90}{2D Original}} &
          \includegraphics[width=0.16\linewidth]{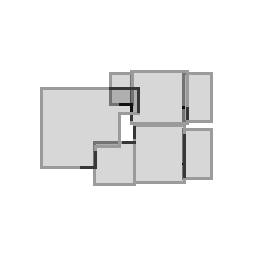} &
          \includegraphics[width=0.16\linewidth]{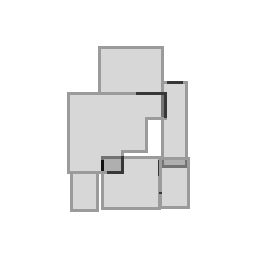} &
          \includegraphics[width=0.16\linewidth]{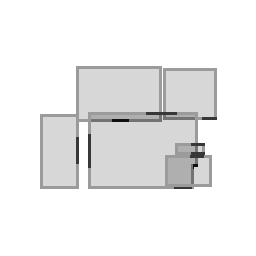} &
          \includegraphics[width=0.16\linewidth]{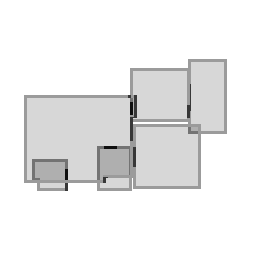} &
          \includegraphics[width=0.16\linewidth]{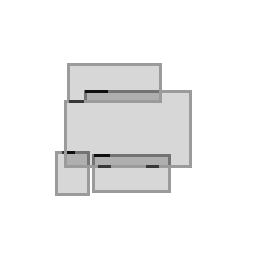} &
          \includegraphics[width=0.16\linewidth]{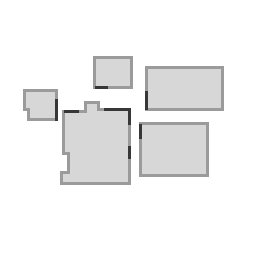} 
          \\[0em]
          \raisebox{2.8em}{\rotatebox{90}{3D Plan}} &
          \includegraphics[width=0.16\linewidth]{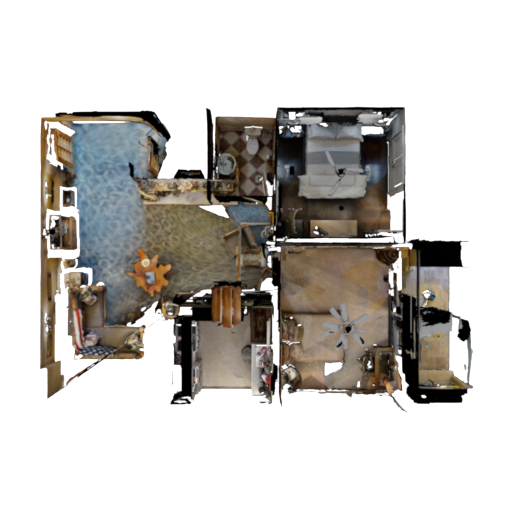} &
          \includegraphics[width=0.16\linewidth]{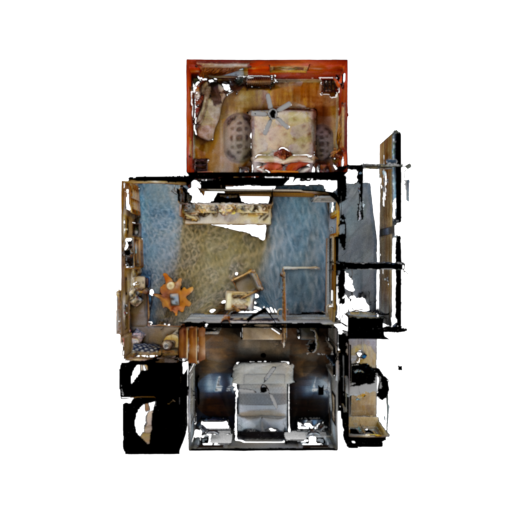} &
          \includegraphics[width=0.16\linewidth]{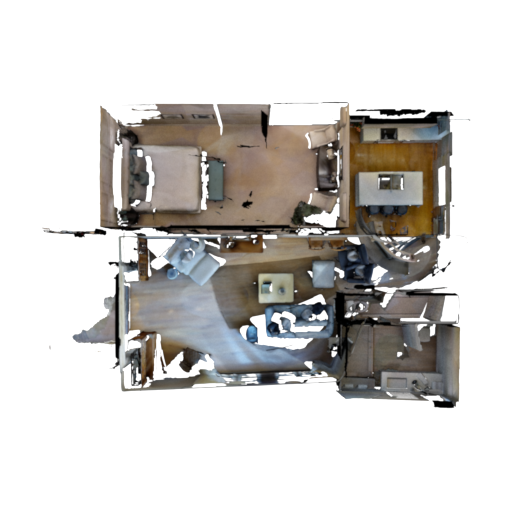} &
          \includegraphics[width=0.16\linewidth]{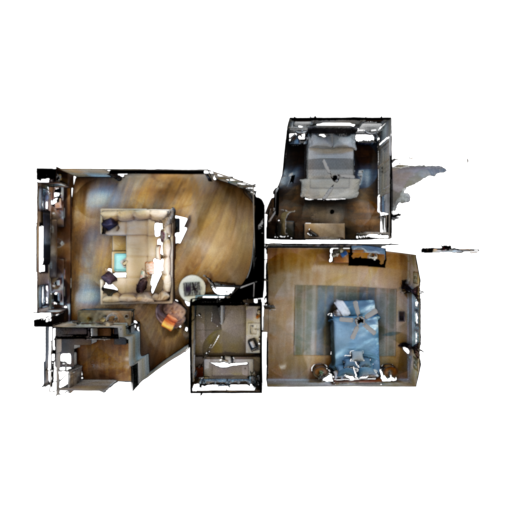} &
          \includegraphics[width=0.16\linewidth]{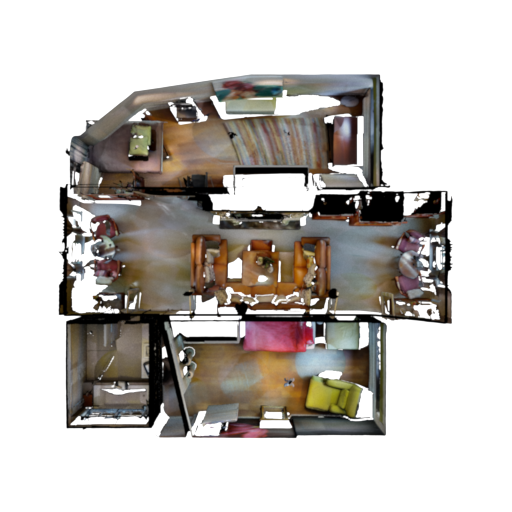} &
          \includegraphics[width=0.16\linewidth]{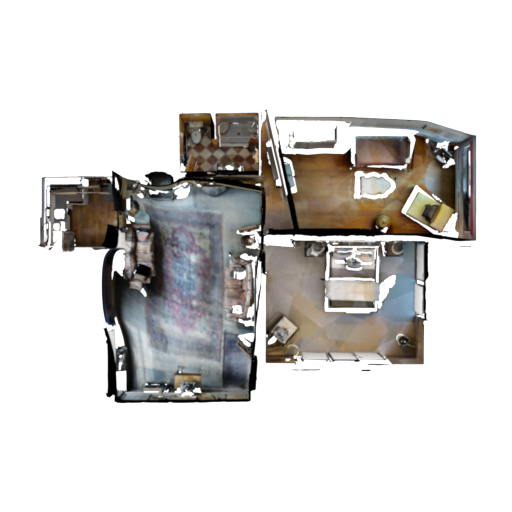}
          \\[3em]
          \multicolumn{7}{c} {Smart Portal Stitching}
          \\[0.5em]
          \raisebox{2.8em}{\rotatebox{90}{2D Plan}} &
          \includegraphics[width=0.16\linewidth]{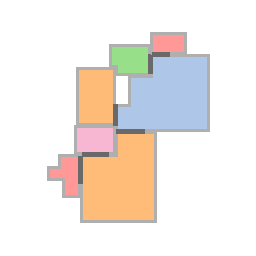} &
          \includegraphics[width=0.16\linewidth]{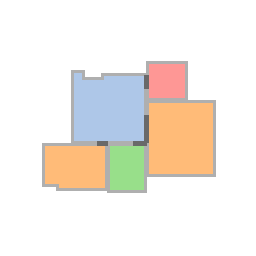} &
          \includegraphics[width=0.16\linewidth]{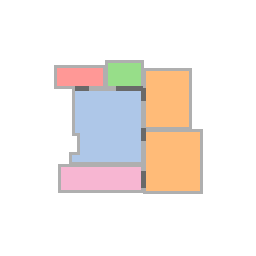} &
          \includegraphics[width=0.16\linewidth]{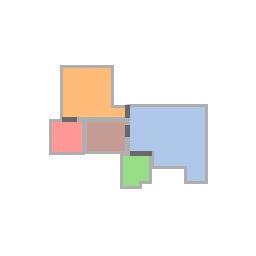} &
          \includegraphics[width=0.16\linewidth]{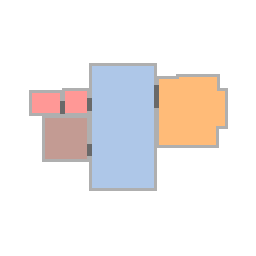} &
          \includegraphics[width=0.16\linewidth]{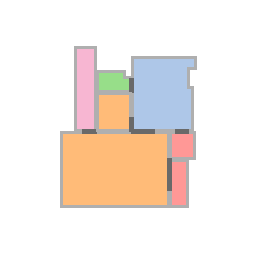}
          \\
          \raisebox{2.25em}{\rotatebox{90}{2D Original}} &
          \includegraphics[width=0.16\linewidth]{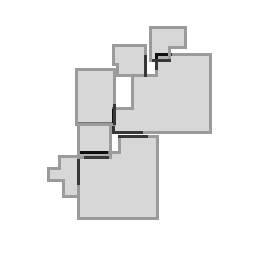} &
          \includegraphics[width=0.16\linewidth]{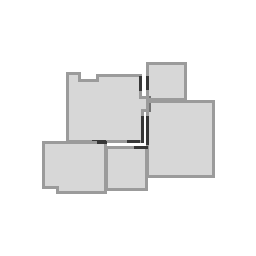} &
          \includegraphics[width=0.16\linewidth]{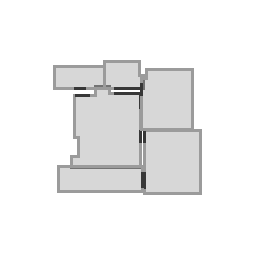} &
          \includegraphics[width=0.16\linewidth]{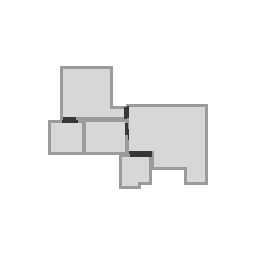} &
          \includegraphics[width=0.16\linewidth]{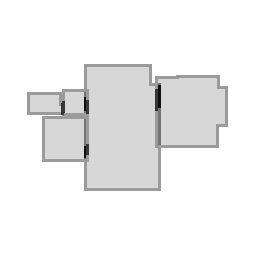} &
          \includegraphics[width=0.16\linewidth]{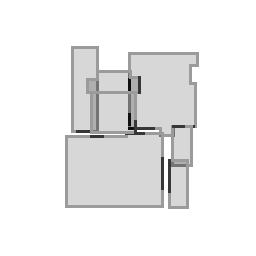} \\[0em]
          \raisebox{2.8em}{\rotatebox{90}{3D Plan}} &
          \includegraphics[width=0.16\linewidth]{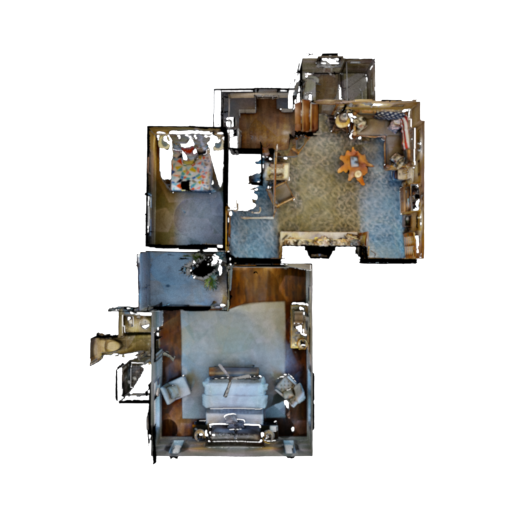} &
          \includegraphics[width=0.16\linewidth]{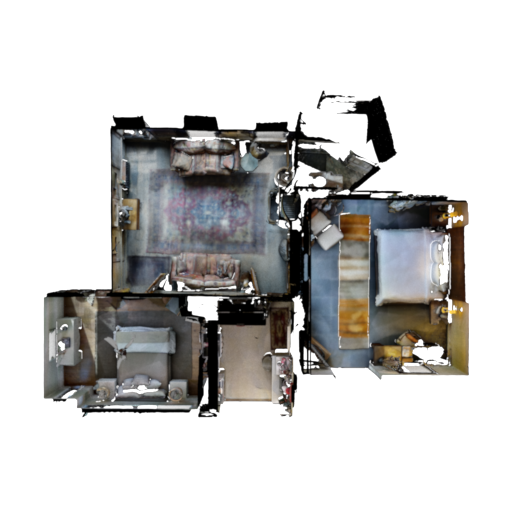} &
          \includegraphics[width=0.16\linewidth]{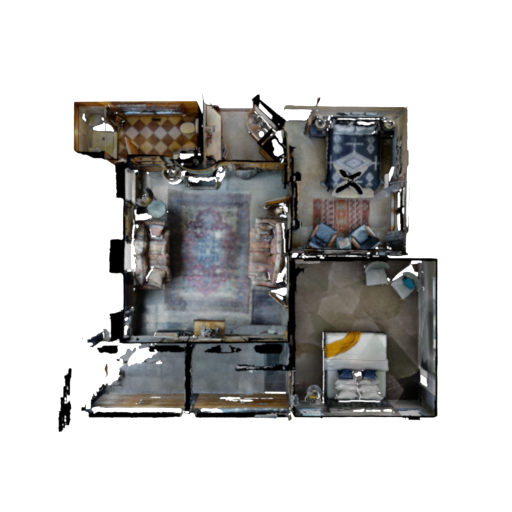} &
          \includegraphics[width=0.16\linewidth]{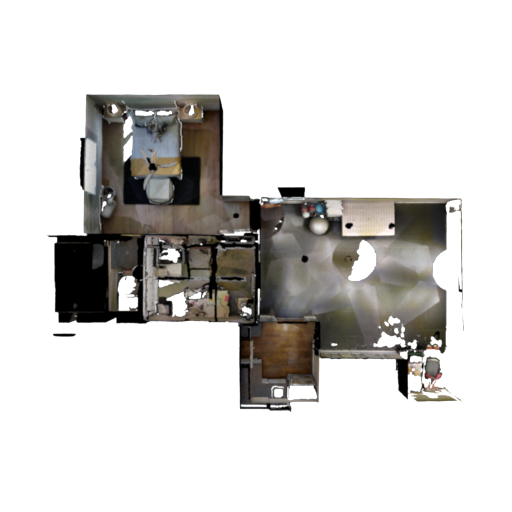} &
          \includegraphics[width=0.16\linewidth]{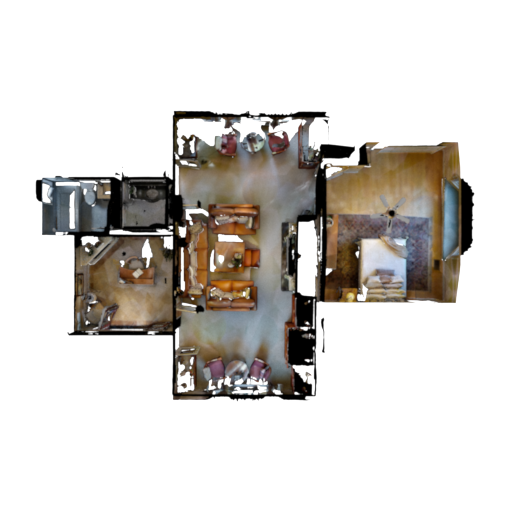} &
          \includegraphics[width=0.16\linewidth]{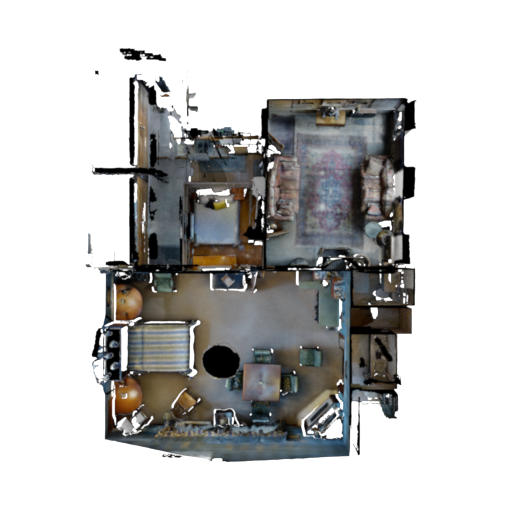}
          \\
    \end{tabular}
    \caption{
    Novel 3D house meshes generated by the two representative strategies, visualized along with the 2D floor plan, and an overlay of the original room shapes and portal location.
    The Match 2D Layout Shape strategy obtains better layouts by directly using existing 2D floor plans.
    However, the 3D rooms do not match the floor plan well, leading to large deformations and visual artifacts.
    In contrast, the Smart Portal Stitching strategy requires minimal deformation to the retrieved 3D room outlines and portals, but at the cost of lower layout quality.
    }
    \label{fig:lots_of_3D_floorplans}
\end{figure*}

Table~\ref{tab:2d_comparisons} shows the results of this experiment.
As expected, sourcing 2D layouts from a floor plan dataset or a data-diven 2D floor plan generative model produces the highest quality layouts.
The smart portal stitching methods sacrifice layout plausibility, although the version with the learned retrieval network performs better.
Figure~\ref{fig:lots_of_3D_floorplans} shows examples of dataset layouts vs. those produced by smart portal stitching.
The layouts generated by the smart portal stitching method have less consistent outlines and sometimes contain implausible features, such as interior holes.
In extreme cases, the layout quality can be too low to be useful, as visualized in Figure~\ref{fig:failurea} and~\ref{fig:failureb}.

\subsection{Evaluating 3D Mesh Quality}

Large deformations of 3D room meshes will degrade perceived visual quality of the output scenes, which is important to most downstream tasks.
We propose the following metrics to evaluate mesh quality in the presence of deformations:
\begin{itemize}
    \item \textbf{Area Change (Area):} Percent change in the area of the room's 2D outline after deformation.
    \item \textbf{Outline Change (Outline):} Average distance between corresponding points on the outlines
    \item \textbf{Portal Change (Portal):} Average distance between the midpoint of the portal after the deformation and the final portal location (obtained through deforming the wall segment)
    \item \textbf{Mesh Statistics}: For each mesh, we discretize the area of each triangle and the length of each edge into 64 bins, and compute the Wasserstein distance of the triangle area distribution (\textbf{Mesh A}) and edge length distribution (\textbf{Mesh E}).
\end{itemize}

\noindent
We use these metrics to compare the following 3D room retrieval methods:
\begin{itemize}
    \item \textbf{Match 2D Layout Shape:} Retrieve rooms whose shape best matches the shape of their corresponding room in a given input 2D floor plan (i.e. the algorithm from Section~\ref{sec:layoutfirst}).
    \item \textbf{Smart Portal Stitching:} Same as above.
    \item \textbf{Smart Portal Stitching (no net):} Same as above.
\end{itemize}

\begin{table}[t!]
    \centering
    \footnotesize
    \setlength{\tabcolsep}{3pt}
    \caption{
    Comparing different methods for retrieving 3D rooms in terms of their effect on final mesh quality.
    Lower is better.
    }
    \begin{tabular}{@{}lccccc@{}}
        \toprule
        \textbf{Method} & \textbf{Area} & \textbf{Outline} & \textbf{Portal} & \textbf{Mesh A} & \textbf{Mesh E} \\
        \midrule
        Match 2D Layout Shape & $16.31$ & $7.776$ & $2.758$ & $4.47$ & $3.19$\\
        Smart Portal Stitching & $10.45$ & $0.026$ & $0.008$ & $2.42$ & $1.83$\\
        Smart Portal Stitching (no net) & $19.32$ & $0.460$ & $0.752$ & $3.47$ & $2.30$\\
        \bottomrule
    \end{tabular}
    \label{tab:3d_comparisons}
\end{table}

\begin{table}[t!]
    \centering
    \caption{
    Evaluating how well a pretrained DDPPO~\cite{wijmans2020ddppo} agent navigates scenes generated from different sources. Higher is better.
    }
    \begin{tabular}{@{}lcc@{}}
        \toprule
        \textbf{Scene Source} & \textbf{Success Rate} & \textbf{SPL} \\
        \midrule
        Match 2D Layout Shape & $0.783$ & $0.608$\\
        Smart Portal Stitching & $0.759$ & $0.618$ \\
        MP3D Scenes & $0.876$ & $0.785$ \\
        \bottomrule
    \end{tabular}
    \label{tab:nav_comparisons}
\end{table}

\begin{table}[t!]
    \centering
    \caption{
    \revisionnew{
    Navigation agent performance for agents trained on scenes generated using our approach and original Matterport3D dataset scenes. Higher values are better. Evaluation is done on the Gibson~\cite{GibsonEnv} dataset, in scenes unseen at training time.
    }
    }
    \begin{tabular}{@{}lcc@{}}
        \toprule
        \textbf{Scene Source} & \textbf{Success Rate} & \textbf{SPL} \\
        \midrule
        Smart Portal Stitching & $0.753$ & $0.484$ \\
        MP3D Scenes & $0.818$ & $0.628$ \\
        \bottomrule
    \end{tabular}
    \label{tab:nav_training}
\end{table}

\begin{figure}[t!]
    \begin{subfigure}[b]{0.24\linewidth}
        \frame{\includegraphics[width=\linewidth]{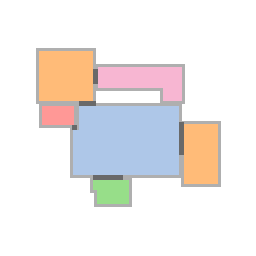}}
        \caption{}
        \label{fig:failurea}
    \end{subfigure}
    \begin{subfigure}[b]{0.24\linewidth}
        \frame{\includegraphics[width=\linewidth]{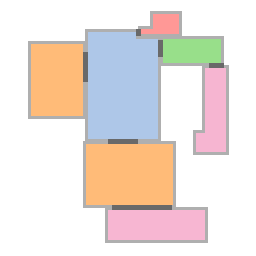}}
        \caption{}
        \label{fig:failureb}
    \end{subfigure}
    \begin{subfigure}[b]{0.24\linewidth}
        \frame{\includegraphics[width=\linewidth]{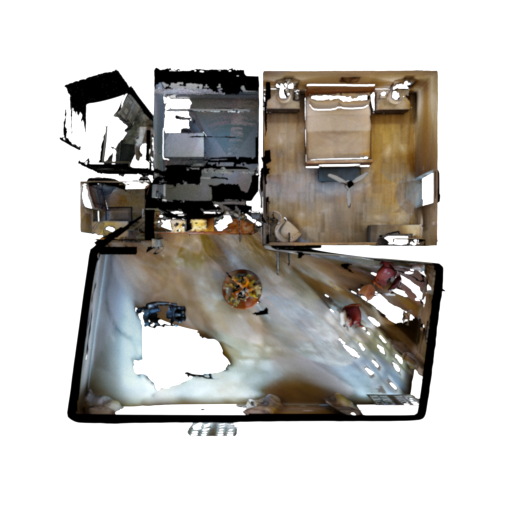}}
        \caption{}
        \label{fig:failurec}
    \end{subfigure}
    \begin{subfigure}[b]{0.24\linewidth}
        \frame{\includegraphics[width=\linewidth]{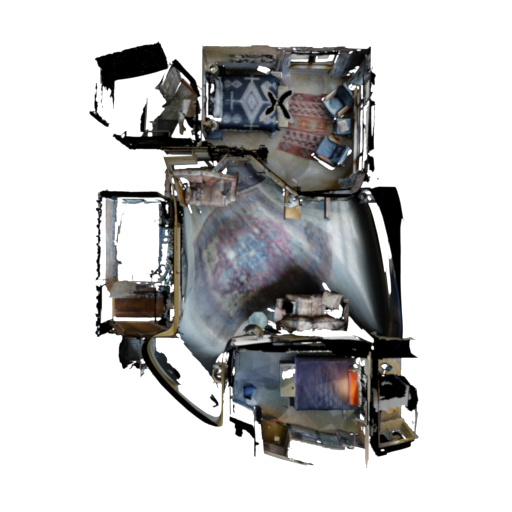}}
        \caption{}
        \label{fig:failured}
    \end{subfigure}
\centering
\setlength{\tabcolsep}{2.5pt}
\caption{Failure cases of different algorithms. (a, b): the Smart Portal Stitching strategy generates a low quality 2D layout that contains large holes or non-smooth outlines; (c, d): the Match and Deform strategy fails to find reasonably similar living rooms, leading  to large deformations in the final mesh.}
\label{fig:failure_cases}
\vspace{-1em}
\end{figure}

Table~\ref{tab:3d_comparisons} shows the results of this experiment. 
The smart portal stitching strategy requires significantly fewer changes to the retrieval 3D rooms, with respect to the size, outline shape, portal locations, and mesh properties.
Using a neural network further amplifies this effect.
On the other hand, due to the limited availability of 3D rooms, the match and deform strategy struggles to find 3D rooms that match exactly, and thus has to modify each room much more.
Figure~\ref{fig:lots_of_3D_floorplans} illustrates the differences in terms of amount of change required and impact on the 3D mesh quality.
In extreme cases, the retrieved rooms are so incompatible that leads to unacceptable mesh quality post deformation. Figure~\ref{fig:failurec} and~\ref{fig:failured} shows two such examples.

\subsection{Evaluating on Downstream Tasks}
To demonstrate that the 3D house scenes we generate through \taskname can be useful for downstream tasks, we use them for \revisionnew{two \revisiontwo{small-scale} indoor visual navigation experiments.}
For the first experiment, we use a DDPPO agent~\cite{wijmans2020ddppo} trained on the Gibson~\cite{GibsonEnv} dataset in the Habitat simulation environment~\cite{AiHabitat}.
We task the agent with performing a series of point goal navigation tasks in scenes generated by our method, as well as in original Matterport3D scenes.
To evaluate how well the agent navigates these scenes, we use two standard metrics from the visual navigation literature: Success Rate and Success Weighted by Path Length (SPL)~\cite{EmbodiedNavEval}.
\revisiontwo{
The pre-trained agent can navigate reasonably well in the scenes generated by both methods.}
Additional qualitative examples of agents navigating in the generated scenes are in the supplementary material.
\revisionnew{
For the second experiment, we train two agents with proximal policy optimization~\cite{PPO} with the settings used in~\cite{AiHabitat}, one on 104 floor plans generated by the Smart Portal Stitching strategy, and the other on 184 floor plans from Matterport3D.
Both agents are trained with 30 million steps.
We then evaluate the performance of the agents on the Gibson~\cite{GibsonEnv} validation set as used in the Habitat challenge~\cite{AiHabitat}.
Table~\ref{tab:nav_training} summarizes the results.
Agents trained on our data achieve comparable success rate to the agent trained on the original Matterport3D data, though with relatively lower SPL.
}
\revisiontwo{
These experiments suggest that data generated by our method is of comparable quality to original Matterport3D data with respect to indoor visual navigation. 
A full-scale evaluation of the impact of data augmentation with our data is computationally challenging, as it has been shown that navigation agents continue to learn from data after billions of steps~\cite{wijmans2020ddppo}. 
The supplemental material reports on an experiment intended to approximate this setting; we leave rigorous, full-scale evaluation to future work.
Regardless, it has been shown that additional training data is beneficial to the performance of agents, even if the additional data is of poor quality~\cite{wijmans2020ddppo}. Thus, we believe that the scenes we generate, which we have shown to be of reasonable quality, can be used to improve the performance of navigation agents further.
}

\revisionnew{
\subsection{Timing}
It takes on average about 30 seconds for the Match 2D Layout Shape strategy to complete a 2D layout. 
The Smart Portal Stitching strategy, in contrast, requires about 10 minutes per floor plan. Virtually all of the time is spent on 2D layout optimization, where per-room-insertion optimization time varies considerably depending upon layout complexity (from 0.1 to 60 seconds, which is our hard time limit).
We note that the time spent on optimization can be reduced from 10 minutes down to 1 minute without beam search.
Computing outline correspondence takes about 1 minute per floor plan with 250 point samples.
This can be reduced drastically if done with fewer point samples. 
Finally, deforming the 3D mesh takes 3 minutes per floor plan, primarily due to the large number of vertices we have to handle.
}

%% file: sections/8-conclusion.tex
\section{Conclusion}
\label{sec:conclusion}

In this paper, we presented \taskname, a new task for creating house-level 3D environments by mixing-and-matching existing 3D rooms. We examined the possible solutions to three sub-tasks, and implemented two representative algorithms for solving the task.
As discussed in Section~\ref{sec:evaluation}, we found \taskname to be challenging, primarily due to the need to trade-off between 2D layout quality and 3D mesh quality.
There are several potential directions one can take to address such challenges.
First, the 2D floor plan quality for the ``2D by 3D'' class of methods can be improved by designing additional semantic features capturing what constitutes a good floor plan.
\revisionnew{For example, one could attempt to reduce the number of corners of the generated 2D floor plans to make them more regular, or fine-tune the shape of individual rooms to make the floor plan more similar to real ones.}
The 3D room deformation procedure can also be improved.
Our current method does not fare well in the case of large deformations, particularly when portals need to slide significantly along walls.
Future work might instead perform additional mesh surgery e.g. cutting and pasting the portal regions to a different part of the wall, instead of deforming the entire wall, \revisionnew{or copying and pasting some existing regions, instead of applying large deformations to those regions.}
Doing so leads to the need to synthesize new textures, which is a challenging problem on its own.
\revisionnew{
Currently, there is no guarantee that the layout of the 3D room post-deformation is still realistic.
Thus, it would be beneficial to augment the control points with semantic layout rules learned from available 2D layout data.
This can be extended further to take into account relations between individual rooms.
}
It is possible to experiment with additional 3D room datasets such as ScanNet~\cite{ScanNet} or Gibson~\cite{GibsonEnv}, in order to increase the chance of retrieving rooms that fit the target floor plan well.
\revisiontwo{
Finally, a more thorough evaluation in downstream tasks such as visual navigation, as well as other tasks that benefit from 3D scene datasets can reveal what properties of the generated scenes are most beneficial in such tasks.
}

Beyond addressing these limitations, there are several broader avenues for future work.
For instance, one can extend \taskname to support on-demand data generation, creating scenes tailored to address scenarios where methods for downstream tasks are under-performing. 
It is also worth doing more investigation on strategies that bridge the gap between different datasets: in this work, we simply combined one dataset of 2D floor plans with another dataset of 3D rooms.
While the performance is acceptable, a more careful investigation addressing differences between datasets can potentially improve the generalization performance of the algorithms.

Finally, we would be excited to see Roominoes applied at scale to more of the applications that originally motivated its creation: computer vision, scene understanding, 3D reconstruction, and embodied AI.
There are a number of interesting questions for future work.
How do we find the right trade-off between 2D and 3D quality depending on the type of downstream tasks?
Does training models for these applications on massive sets of Roominoes-generated 3D floor plans improve their generalization performance?
And is it possible to reduce the number of training scenes needed if those scenes are tailored to the task(s) the method must address?
These are some exciting questions that would be fruitful to explore with Roominoes.

%% file: sections/9-acknowledgements.tex
\section*{Acknowledgments}

We thank Supriya Gadi Patil and Jingling Li for discussions in the early stage of the project, and the anonymous reviewers for their helpful suggestions.
An earlier version of this project was started when Kai Wang interned at Facebook AI Research, with helpful inputs from Dhruv Batra, Oleksandr Maksymets and Erik Wijimans.
This work was funded in part by NSF award \#1907547.
Angel X. Chang is supported by a Canada CIFAR AI Chair and NSERC Discovery Grant, and Manolis Savva by a Canada Research Chair and NSERC Discovery Grant.
Daniel Ritchie is an advisor to Geopipe and owns equity in the company.
Geopipe is a start-up that is developing 3D technology to build immersive virtual copies of the real world with applications in various fields, including games and architecture.

%% file: supplementary_sections/1-optimization.tex
\section{Optimizing Post-Retrieval Layouts}
\label{sec:MIQP}
In this section, we describe the details optimization process we take to align the rooms retrieved by the retrieval process described in Section~5 of the main paper.
Inspired by prior work~\cite{MIQPLayout,GenerativeLayoutModelingConstraintGraphs}, we adopt a mixed integer quadratic programming (MIQP) based procedure for layout optimization.

\subsection{Decomposing Rooms into Rectangles}
Rooms in our representation are arbitrary rectilinear polygons.
It is intractable to define certain important constraints, such as non-overlap, between such polygons.
Thus, our first step is to decompose each room into a set of rectangles.
We then use constraints to bind these rectangles together so they behave as a continuous room.

\begin{figure}[t!]
    \centering
    \setlength{\tabcolsep}{0pt}
    \renewcommand{\arraystretch}{0}
    \begin{tabular}{cc}
        Original Rooms & Decomposed Rectangles\\
        \includegraphics[width=.49\linewidth]{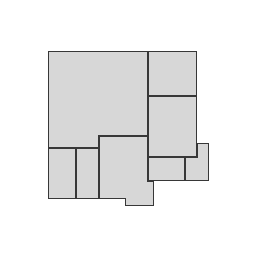} &
        \includegraphics[width=.49\linewidth]{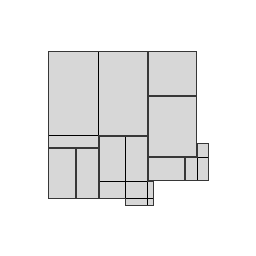}
    \end{tabular}
    \caption{Example of a floor plan before and after rectangle decomposition}
    \label{fig:decomposition}
\end{figure}

In general, the decomposition of a rectilinear polygon into rectangles is not unique.
In our implementation, we use the \emph{maximal} decomposition, which is found by constructing a grid over all vertex coordinates and then taking the grid cells which fall inside the polygon as the decomposed rectangles (Figure~\ref{fig:decomposition}).
The maximal decomposition is quick to compute and also has the property that every pair of adjacent rectangles shares a complete edge.
This edge-sharing property makes it easy to impose additional constraints to bind the decomposed rectangles of a given room together.

\subsection{Optimization Variables}

In our solver, we express each rectangle $i$ in terms of four variables: its upper-left vertex position $\langle x_i, y_i \rangle$, its width $w_i$, and its height $h_i$.
In addition to the room shapes, we must also represent and optimize for the positions and sizes of portals between rooms (e.g. doors), so that the resulting floor plan is navigable.
We describe a portal $j$ as a line segment with centroid position $\langle px_j, py_j\rangle$ and radius (i.e. half-length) $pr_j$.
Each portal is permitted to slide along certain walls of the room which contains it, as the next section will describe.

\subsection{Constraints}

\subsubsection*{Non-negativity} All variables must be non-negative so that our solution lies in the positive quadrant and no output rectangles or portals have a (non-physical) negative width or height:
\begin{equation*}
x_i, y_i, w_i, h_i, px_j, py_j, pr_j \geq 0 \qquad \forall i, j
\end{equation*}

\subsubsection*{Minimal Room Size}
To prevent the optimization from collapsing certain rooms in favor of others, we enforce that each room has a width and height of at least $s$.
% \changes{Find what that value is}.
To do this, we identify a sequence of indices $R^x$ of rectangles that spans the horizontal extent of the room, as well as a sequence of  $R^y$ for rectangles that spans its vertical extent. Then:
\begin{equation*}
\sum_{i=0}^{|R^x|} w_{R^x_i} > s , \sum_{i=0}^{|R^y|} h_{R^y_i} > s  \qquad \forall i \in r  
\end{equation*}

\subsubsection*{Non-overlap} We require the solution to have no overlapping pairs of rectangles $i, j$. There are four possible relationships to account for: $i$ is either to the top, bottom, left, or right side of $j$. Let $D \in \{T, B, L, R\}$ represent these relationships respectively. 
We let the MIQP optimizer select from this set of possible relationships by introducing an auxiliary binary variable $\sigma_{i, j}^{D}$,
where $\sigma_{i, j}^D = 1$ if and only rectangles $i$ and $j$ have the relationship $D$. This results in the following set of constraints for each pair of rectangles $i, j$:
\begin{align*}
        x_i - w_j &\geq x_j - M\cdot(1 - \sigma_{i, j}^R)\\
        x_i + w_i &\leq x_j + M\cdot(1 - \sigma_{i, j}^L)\\
        y_i - h_j &\geq y_j - M\cdot(1 - \sigma_{i, j}^B)\\
        y_i + d_i &\leq y_j + M\cdot(1 - \sigma_{i, j}^T)\\
        \sum_{D=1}^4 \sigma_{i, j}^D &\geq 1
\end{align*}
where $M$ is a large constant to ensure that rectangles $i$ and $j$ do not overlap in direction $D$ when $\sigma_{i, j}^D = 1$ (we set $M = x_\text{max} + y_\text{max}$ in our implementation). The last constraint requires that at least one of the four auxiliary variables has a value of 1.

\subsubsection*{Decomposition constraints}
For each decomposed room, we introduce the following constraints for all pairs of its rectangles $i, j$ such that $i$ is to the left of $j$ (the rectangles share a vertical edge):
\begin{equation*}
    x_i + w_i = x_j \qquad y_i = y_j \qquad h_i = h_j
\end{equation*}
and the following constraints for all pairs of rectangles $i, j$ such that $i$ is to the top of $j$ (the rectangles share a horizontal edge):
\begin{equation*}
    y_i + h_i = y_j \qquad x_i = x_j \qquad w_i = w_j
\end{equation*}
These constraints bind the rectangles together such that they maintain shared edges.

\subsubsection*{Portal connection}
If two portals $i, j$ are specified as connected, then their positions and half-lengths must be equivalent:
\begin{equation*}
    px_i = px_j \qquad py_i = py_j \qquad pr_i = pr_j
\end{equation*}

\subsubsection*{Portal sliding} 
We require that portals stay on the same wall that they are initially defined to be on, and that their position and length do not extend beyond this wall.
If a room decomposes to a single rectangle, then the portal can slide along one edge $D$ of this rectangle (for example, $D = T$ means the portal lies on the top wall).
The sliding constraint thus takes on one of four cases:
\begin{align*}
    &\text{ if D = T } 
    \begin{cases}
        py_j = y_i\\
        px_j \geq x_i + pr_j\\
        px_j \leq x_i + w_i - pr_j
    \end{cases}
    &\text{ if D = B }
    \begin{cases}
        py_j = y_i + h_i\\ 
        px_j \geq x_i + pr_j\\
        px_j \leq x_i + w_i - pr_j 
    \end{cases}
    \\
    &\text{ if D = L }
    \begin{cases}
        px_j = x_i \\ 
        py_j \geq y_i + pr_j\\
        py_j \leq y_i + h_i - pr_j
    \end{cases}
    &\text{ if D = R }
    \begin{cases}
        px_j = x_i + w_i \\ 
        py_j \geq y_i + pr_j\\
        py_j \leq y_i + h_i - pr_j
    \end{cases}
\end{align*}
In general, a room decomposes into multiple rectangles.
Here, we must handle the case where a portal lies on a room wall that is shared by more than one decomposed rectangle.
This scenario uses the same form of constraint as above, but requires us to know the indices of the rooms between which the portal can slide.
For instance, if a portal $l$ slides along the left side of a wall shared by three rectangles $i, j, k$ where $i$ is the top-most rectangle and $k$ is the bottom-most, then the constraints would be:
\begin{equation*}
        px_l = x_i \qquad py_l \geq y_i + pr_l \qquad py_l \leq y_k + h_k - pr_l 
\end{equation*}

\subsection{Objective}

There may be multiple floor plan configurations which satisfy all the constraints defined above.
Within this feasible set, there are certain configurations which are preferable.
Primarily, we prefer layouts that change room shapes and portal positions/sizes as little as possible, as such changes will introduce distortion when transferred to the 3D mesh.

Secondarily, we prefer layouts which maximize the number of wall-to-wall adjacencies between rooms, as this results in more plausibly compact/space-efficient layouts and also avoids introducing interior voids in the layout.
To keep track of adjacenciies, we use a similar formulation to the non-overlap constraint.
We add a binary variable $\sigma^A_{i, j}$ for every pair of rectangles $(i,j)$ to indicate whether the two rectangles should be adjacent, and then we add the following constraints to account for possible adjacency relationships:
\begin{equation*}
    \begin{cases}
        x_i \leq x_j + w_j - L\cdot\theta_{i, j} + M\cdot(1 - \sigma^A_{i, j}) \\
        x_i + w_i \geq x_j + L\cdot\theta_{i, j} - M\cdot(1 - \sigma^A_{i, j}) \\
        y_i \leq y_j + h_j - L\cdot(1 - \theta_{i, j}) + M\cdot(1 - \sigma^A_{i, j}) \\
        y_i + h_i \geq y_j + L\cdot(1 - \theta_{i, j}) - M\cdot(1 - \sigma^A_{i, j})\\
    \end{cases}
\end{equation*}
where $\theta_{i, j}$ is a binary variable for whether the rectangles are horizontally or vertically adjacent, and $L$ is the minimum length of the line segment $i$ and $j$ must share to be considered adjacent (we use $L = 6$).

Finally, we minimize the following overall objective function:
\begin{align*}
    &\lambda_{1} (\norm{\mathbf{w} - \mathbf{\hat{w}}}^2 + \norm{\mathbf{h} - \mathbf{\hat{h}}}^2) + 
    \lambda_{2} \norm{\mathbf{pr} - \mathbf{\hat{pr}}}^2 - \lambda_{3} \sum_{i,j} \sigma^A_{i, j} \cdot \indicator(i,j)
    \\
    &+ \lambda_{4} \sum_{i \in P_\text{vert}} ( (py_i - y_{T_i}) - (\hat{py}_i - \hat{y}_{T_i}) )^2 + ( (py_i - y_{B_i}) - (\hat{py}_i - \hat{y}_{B_i}) )^2
    \\
    &+ \lambda_{4} \sum_{i \in P_\text{horz}} ( (px_i - x_{L_i}) - (\hat{px}_i - \hat{x}_{L_i}) )^2 + ( (px_i - x_{R_i}) - (\hat{px}_i - \hat{x}_{R_i}) )^2
\end{align*}
where the $\hat{x}$ version of a variable $x$ denotes its initial value.
The first term penalizes changes in room rectangle shape; the second penalizes changes in portal radius.
The third term rewards pairs of adjacent rectangles from different rooms, but only if the rooms containing those two rectangles $i,j$ have already had all rooms marked adjacent to them in the input graph placed into the layout (this is the role of the indicator function $\indicator(i,j)$).
Finally, the last two terms penalize deviations in all portals' positions along their respective walls.
Here, $P_\text{vert}$ and $P_\text{horz}$ return the indices of all vertical and horizontal portals, respectively; $T_i, B_i, L_i$, and $R_i$ give the index of the top, bottom, left, and right adjacent rectangle to portal $i$'s wall, respectively.
In our implementation, we use $\lambda_{1}=1, \lambda_{2}=5, \lambda_{3} = 100, \lambda_{4}=3$, with all rooms scaled with a ratio of $18$ meters to $256$ units.

%% file: supplementary_sections/2-navigation.tex
\section{Qualitative Results for the Navigation Experiment}
\label{sec:nav}
\begin{figure*}[t!]
    \centering
    \setlength{\tabcolsep}{1pt}
    \begin{tabular}{cccccccc}
        \multicolumn{8}{c}{Smart Portal Stitching} \\
        \includegraphics[width=.115\linewidth]{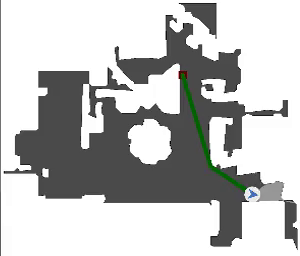} &
        \includegraphics[width=.115\linewidth]{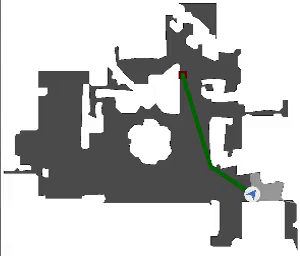} &
        \includegraphics[width=.115\linewidth]{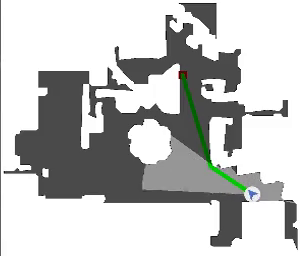} &
        \includegraphics[width=.115\linewidth]{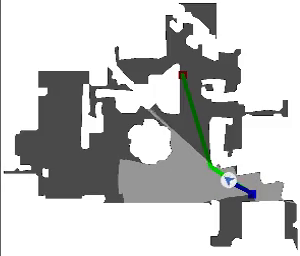} &
        \includegraphics[width=.115\linewidth]{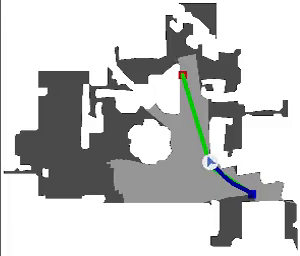} &
        \includegraphics[width=.115\linewidth]{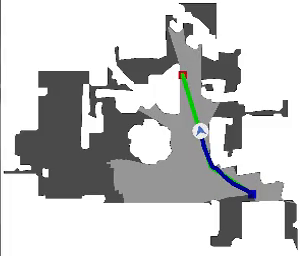} &
        \includegraphics[width=.115\linewidth]{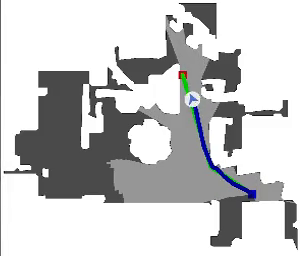} &
        \includegraphics[width=.115\linewidth]{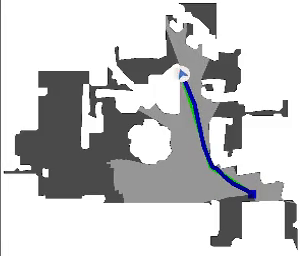}
        \\
        \includegraphics[width=.115\linewidth]{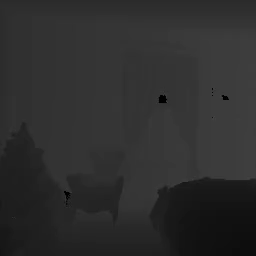} &
        \includegraphics[width=.115\linewidth]{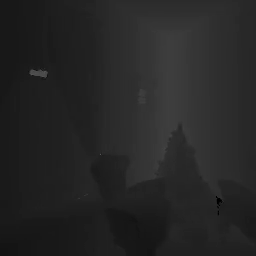} &
        \includegraphics[width=.115\linewidth]{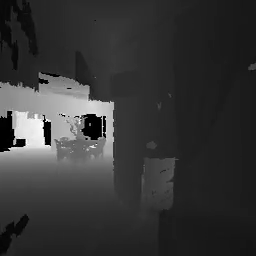} &
        \includegraphics[width=.115\linewidth]{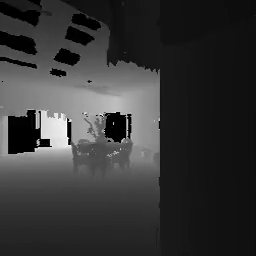} &
        \includegraphics[width=.115\linewidth]{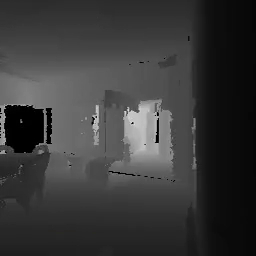} &
        \includegraphics[width=.115\linewidth]{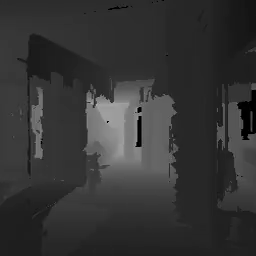} &
        \includegraphics[width=.115\linewidth]{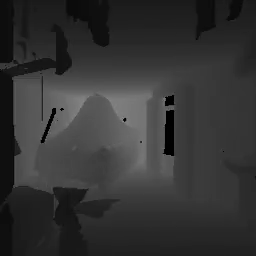} &
        \includegraphics[width=.115\linewidth]{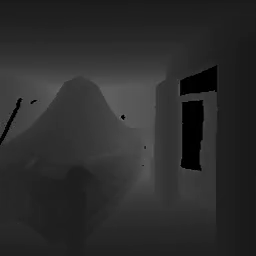}
        \\
        \midrule
        \multicolumn{8}{c}{Match 2D Layout Shape} \\
        \includegraphics[width=.115\linewidth]{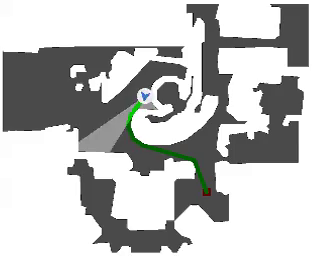} &
        \includegraphics[width=.115\linewidth]{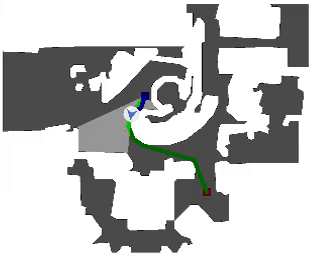} &
        \includegraphics[width=.115\linewidth]{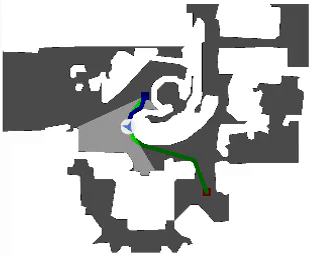} &
        \includegraphics[width=.115\linewidth]{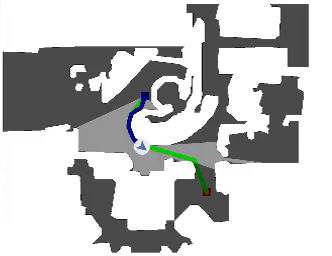} &
        \includegraphics[width=.115\linewidth]{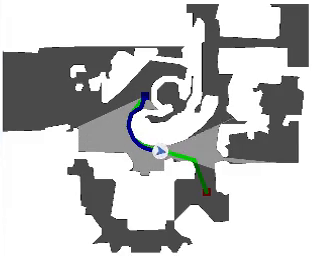} &
        \includegraphics[width=.115\linewidth]{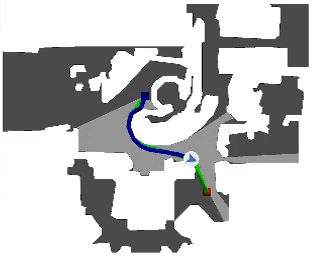} &
        \includegraphics[width=.115\linewidth]{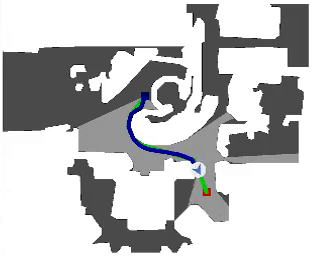} &
        \includegraphics[width=.115\linewidth]{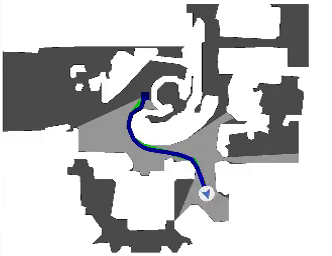}
        \\
        \includegraphics[width=.115\linewidth]{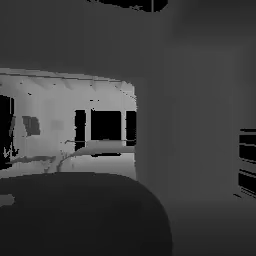} &
        \includegraphics[width=.115\linewidth]{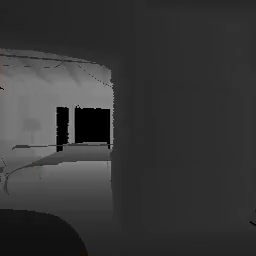} &
        \includegraphics[width=.115\linewidth]{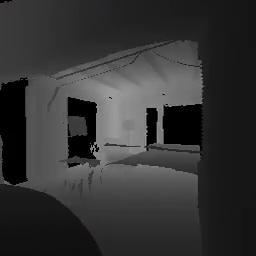} &
        \includegraphics[width=.115\linewidth]{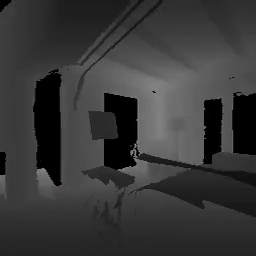} &
        \includegraphics[width=.115\linewidth]{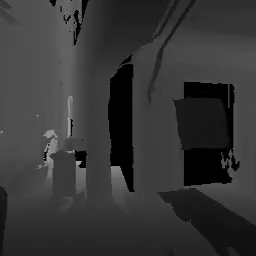} &
        \includegraphics[width=.115\linewidth]{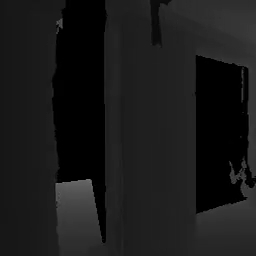} &
        \includegraphics[width=.115\linewidth]{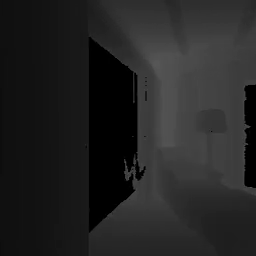} &
        \includegraphics[width=.115\linewidth]{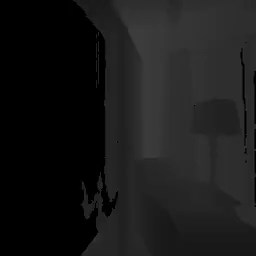}
        \\
        \midrule
        \multicolumn{8}{c}{Matterport 3D} \\
        \includegraphics[width=.115\linewidth]{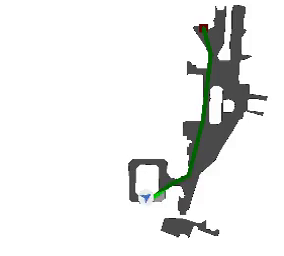} &
        \includegraphics[width=.115\linewidth]{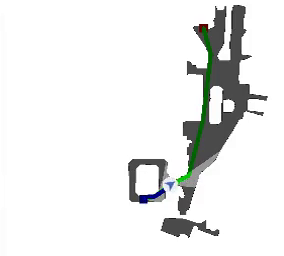} &
        \includegraphics[width=.115\linewidth]{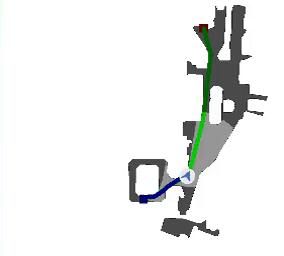} &
        \includegraphics[width=.115\linewidth]{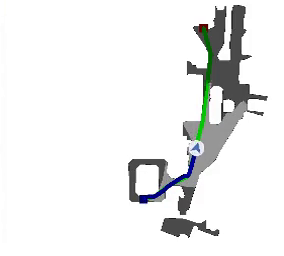} &
        \includegraphics[width=.115\linewidth]{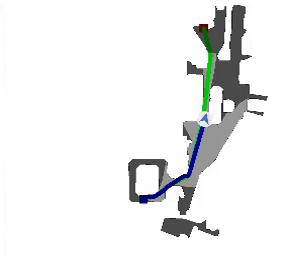} &
        \includegraphics[width=.115\linewidth]{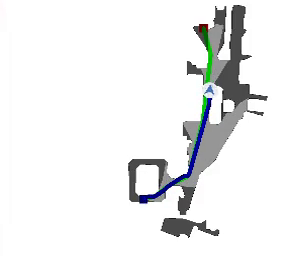} &
        \includegraphics[width=.115\linewidth]{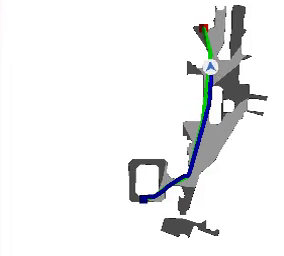} &
        \includegraphics[width=.115\linewidth]{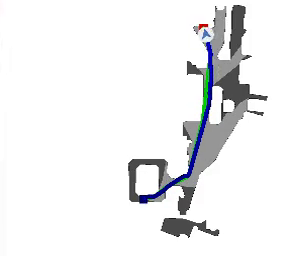}
        \\
        \includegraphics[width=.115\linewidth]{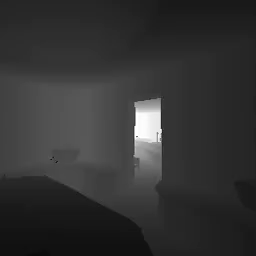} &
        \includegraphics[width=.115\linewidth]{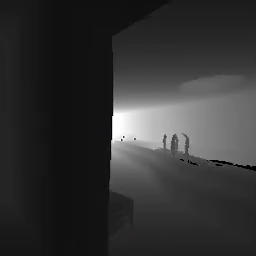} &
        \includegraphics[width=.115\linewidth]{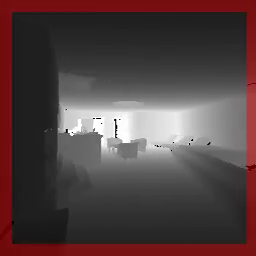} &
        \includegraphics[width=.115\linewidth]{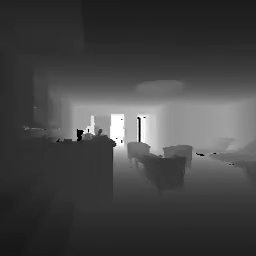} &
        \includegraphics[width=.115\linewidth]{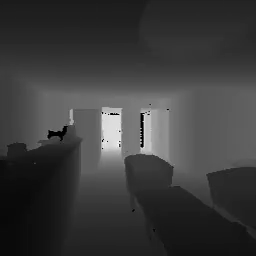} &
        \includegraphics[width=.115\linewidth]{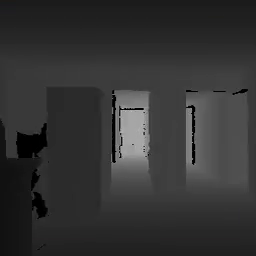} &
        \includegraphics[width=.115\linewidth]{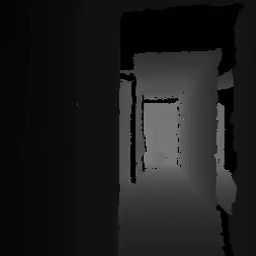} &
        \includegraphics[width=.115\linewidth]{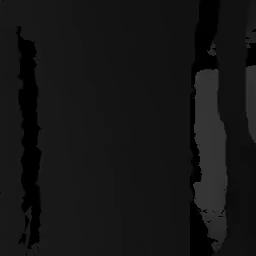}
    \end{tabular}
    \caption{
    Trajectory taken by a DDPPO agent through scenes generated by the proposed methods, as well as a scene taken directly from Matterport3D.
    Top row: top down view of the navigable areas.
    Bottom row: agent's first-person view of the scene, depth only.
    }
    \label{fig:habitat_walkthrough}
\end{figure*}

Figure~\ref{fig:habitat_walkthrough} shows example trajectories of a pre-trained DDPPO~\cite{wijmans2020ddppo} agent walking through scenes generated by our methods, as well an original Matterport3D~\cite{Matterport3D} scene.
The full videos can be found at DDPPO\_portal\_stitching.mp4, DDPPO\_match\_2d.mp4, DDPPO\_mp3d.mp4 respectively. \revisiontwo{In these videos, the left side shows the depth image that the agent is seeing, whereas the right side shows the navigable areas of the scene, with the shortest trajectory to goal visualized in green and the trajectory the agent took visualized in blue.}

%% file: supplementary_sections/3-walkthrough.tex
\section{Walk-through of a Generated Scene}
\label{sec:walkthrough}

\begin{figure*}[t!]
    \centering
    \setlength{\tabcolsep}{1pt}
    \begin{tabular}{cccc}
        \includegraphics[width=.24\linewidth]{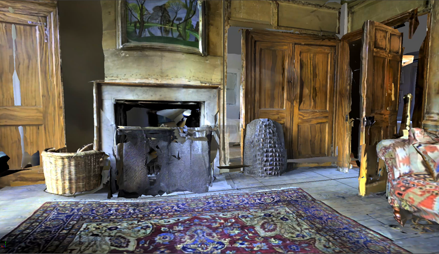} &
        \includegraphics[width=.24\linewidth]{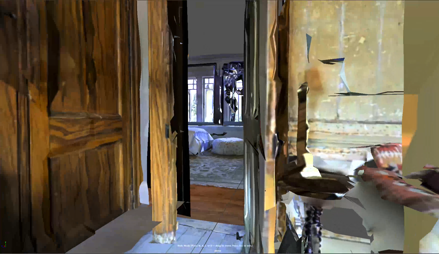} &
        \includegraphics[width=.24\linewidth]{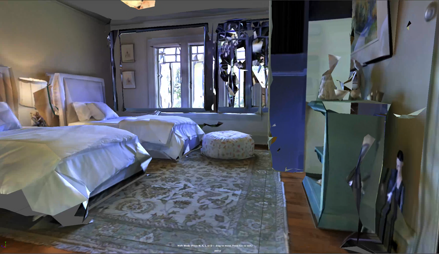} &
        \includegraphics[width=.24\linewidth]{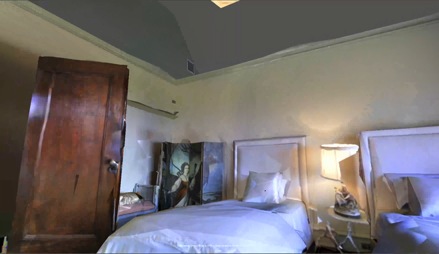}\\
        
        \includegraphics[width=.24\linewidth]{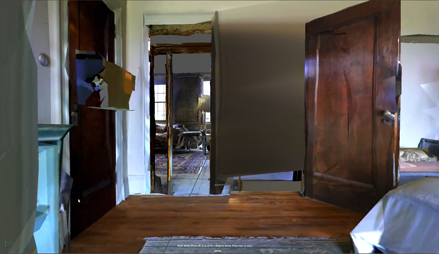} &
        \includegraphics[width=.24\linewidth]{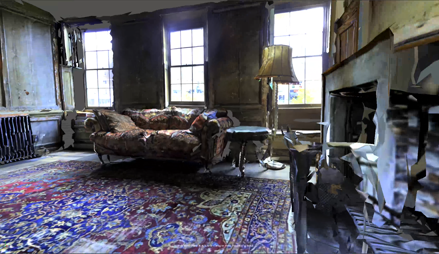} &
        \includegraphics[width=.24\linewidth]{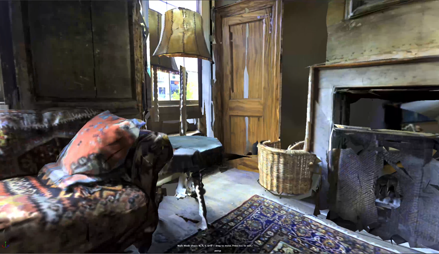} &
        \includegraphics[width=.24\linewidth]{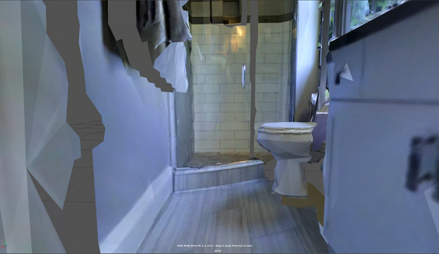}\\
        
        \includegraphics[width=.24\linewidth]{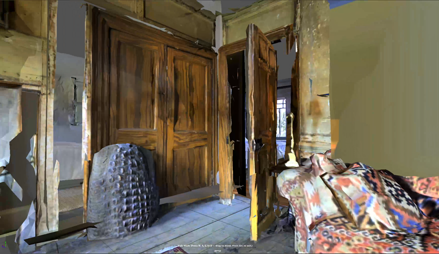} &
        \includegraphics[width=.24\linewidth]{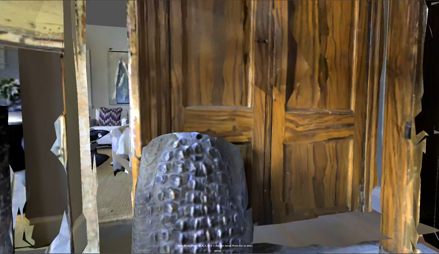} &
        \includegraphics[width=.24\linewidth]{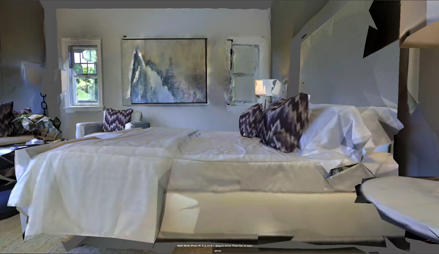} &
        \includegraphics[width=.24\linewidth]{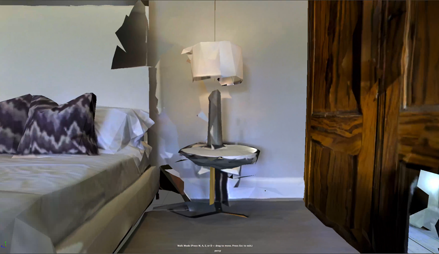}\\
    \end{tabular}
    \caption{
    Trajectory of a first person walk through for a scene generated by the smart portal stitching strategy.
    }
    \label{fig:high_quality_walkthrough}
\end{figure*}

Figure~\ref{fig:high_quality_walkthrough} shows a trajectory of a first-person walk through of one of the generated scenes. The original semantic annotation of Matterpot3D meshes are done on meshes reconstructed with a different pipeline, and subsequently of lower visual quality. To produce this trajectory, we manually annotated the higher quality meshes of the set of rooms contained in a layout generated by the smart portal stitching strategy, \revisiontwo{and then manually performed a walk through.} 
The full video can be found at first\_person\_walkthrough.mp4

%% file: supplementary_sections/4-navigation.tex
\section{Evaluating the Effect of Data Augmentation with Our Data}
\label{sec:nav_sup}

\revisiontwo{
Directly evaluating the impact of data augmentation with our data is challenging, as it has been shown that navigation agents continue to learn from data after billions of steps~\cite{wijmans2020ddppo}.
Here, we provide an approximation by evaluating two agents that perform similarly on their respective training sets.
We train one of the agent on 184 floor plans from the Matterport3D dataset, and the other the 184 Matterport3D floor plans, as well as 104 floor plans generated by the Smart Portal Stitching strategy.
For the second agent, we construct the training set such that half of the episodes are from Matterport3D, and the other half from the generated data.
We train the first agent for 30 million steps. We record the training success rate (about $0.67$) and SPL (about $0.49$), and then train the second agent until it reaches similar performances, at around 50 million steps.
We then evaluate the trained agents on the Gibson validation set. The results are summarized in table~\ref{tab:nav_sup}.
The agent trained on Matterport3D + Smart Portal Stitching outperforms the agent trained on only Matterport3D with respect to both success rate and SPL.
We do stress that this is only an approximation, and it is possible the better performance results from other factors.
We leave rigorous, full-scale evaluation to future works.
}

\begin{table}[t!]
    \centering
    \caption{
    \revisiontwo{
    Evaluating the impact of data augmentation with data generated by our methods. Higher values are better. Evaluation is done on the Gibson~\cite{GibsonEnv} dataset, in scenes unseen at training time.
    }
    }
    \begin{tabular}{@{}lcc@{}}
        \toprule
        \textbf{Scene Source} & \textbf{Success Rate} & \textbf{SPL} \\
        \midrule
        Smart Portal Stitching + MP3D Scenes & $0.831$ & $0.662$ \\
        MP3D Scenes & $0.818$ & $0.628$ \\
        \bottomrule
    \end{tabular}
    \label{tab:nav_sup}
\end{table}